\documentclass{article}

\usepackage{microtype}
\usepackage{graphicx}
\usepackage{subfigure}
\usepackage{booktabs} 

\usepackage{hyperref}

\usepackage[accepted]{icml2023}

\usepackage{amsmath}
\usepackage{amssymb}
\usepackage{mathtools}
\usepackage{amsthm}
\usepackage{adjustbox}

\usepackage{amsmath,amsfonts,bm}

\def\eqref#1{equation~\ref{#1}}

\def\1{\bm{1}}

\def\vt{{\bm{t}}}

\def\ve{{\bm{e}}}

\def\vr{{\bm{r}}}

\def\vt{{\bm{t}}}

\def\vv{{\bm{v}}}

\def\vx{{\bm{x}}}
\def\vy{{\bm{y}}}
\def\vz{{\bm{z}}}

\def\mE{{\bm{E}}}

\def\mH{{\bm{H}}}

\def\mR{{\bm{R}}}

\def\mX{{\bm{X}}}

\DeclareMathAlphabet{\mathsfit}{\encodingdefault}{\sfdefault}{m}{sl}
\SetMathAlphabet{\mathsfit}{bold}{\encodingdefault}{\sfdefault}{bx}{n}

\usepackage{graphicx}
\usepackage{float}
\usepackage{amsfonts,amsmath,amssymb,amsthm}
\usepackage{multicol, multirow}
\usepackage{makecell}
\usepackage{adjustbox}
\usepackage{enumitem}
\usepackage{wrapfig}
\usepackage{algorithm}
\usepackage{algorithmic}
\usepackage{afterpage}

\usepackage[capitalize,noabbrev]{cleveref}

\newcommand{\ie}{\em{i.e.}}
\newcommand{\eg}{\em{e.g.}}

\newcommand{\twoD}{\text{2D}}
\newcommand{\threeD}{\text{3D}}
\newcommand{\localframe}{\text{frame}}

\usepackage{xcolor}
\newcommand{\model}[1]{MoleculeSDE}
\newcommand{\norm}[1]{\left\lVert#1\right\rVert}

\icmltitlerunning{A Group Symmetric Stochastic Differential Equation Model for Molecule Multi-modal Pretraining}

\begin{document}

\twocolumn[

\icmltitle{
A Group Symmetric Stochastic Differential Equation Model\\
for Molecule Multi-modal Pretraining
}

\icmlsetsymbol{equal}{*}

\begin{icmlauthorlist}
\icmlauthor{Shengchao Liu}{equal,mila,udem}
\icmlauthor{Weitao Du}{equal,cas}
\icmlauthor{Zhiming Ma}{cas}
\icmlauthor{Hongyu Guo}{nrc}
\icmlauthor{Jian Tang}{mila,hec}
\end{icmlauthorlist}

\icmlaffiliation{mila}{Mila - Québec AI Institute, Canada}
\icmlaffiliation{udem}{Université de Montréal, Canada}
\icmlaffiliation{cas}{Chinese Academy of Sciences, China}
\icmlaffiliation{nrc}{National Research Council Canada, Canada}
\icmlaffiliation{hec}{HEC Montréal, Canada}

\icmlcorrespondingauthor{Shengchao Liu}{liusheng@mila.quebec}

\icmlkeywords{molecule conformation, geometric pretraining, 2D topology, 3D geometry, group symmetric, SE(3)-equivariant and reflection-antisymmetric, stochastic differential equation (SDE), diffusion model}

\vskip 0.3in
]

\printAffiliationsAndNotice{\icmlEqualContribution}

\begin{abstract}
Molecule pretraining has quickly become the go-to schema to boost the performance of AI-based drug discovery. Naturally, molecules can be represented as 2D topological graphs or 3D geometric point clouds. Although most existing pertaining methods focus on merely the single modality, recent research has shown that maximizing the mutual information (MI) between such two modalities enhances the molecule representation ability. Meanwhile, existing molecule multi-modal pretraining approaches approximate MI based on the representation space encoded from the topology and geometry, thus resulting in the loss of critical structural information of molecules. To address this issue, we propose MoleculeSDE. MoleculeSDE leverages group symmetric ({\eg}, SE(3)-equivariant and reflection-antisymmetric) stochastic differential equation models to generate the 3D geometries from 2D topologies, and vice versa, \textit{directly} in the input space. It not only obtains tighter MI bound but also enables prosperous downstream tasks than the previous work. By comparing with 17 pretraining baselines, we empirically verify that MoleculeSDE can learn an expressive representation with state-of-the-art performance on 26 out of 32 downstream tasks. The source codes are available in \href{https://github.com/chao1224/MoleculeSDE}{this repository}.
\end{abstract}

\section{Introduction} \label{sec:intro}
Artificial intelligence (AI) for drug discovery has recently attracted a surge of research interest in both the machine learning and cheminformatics communities, demonstrating encouraging outcomes in many challenging drug discovery tasks~\cite{gilmer2017neural,gomez2018automatic,liu2019n,corso2020principal,jin2020hierarchical,rampavsek2022recipe,liu2022multi,jumper2021highly,demirel2021attentive,hsu2022learning,liu2023text,yao2023leveraging}. These successes are primarily attributed to the informative representations of molecules. 

\begin{figure*}[t]
\centering
\includegraphics[width=\linewidth]{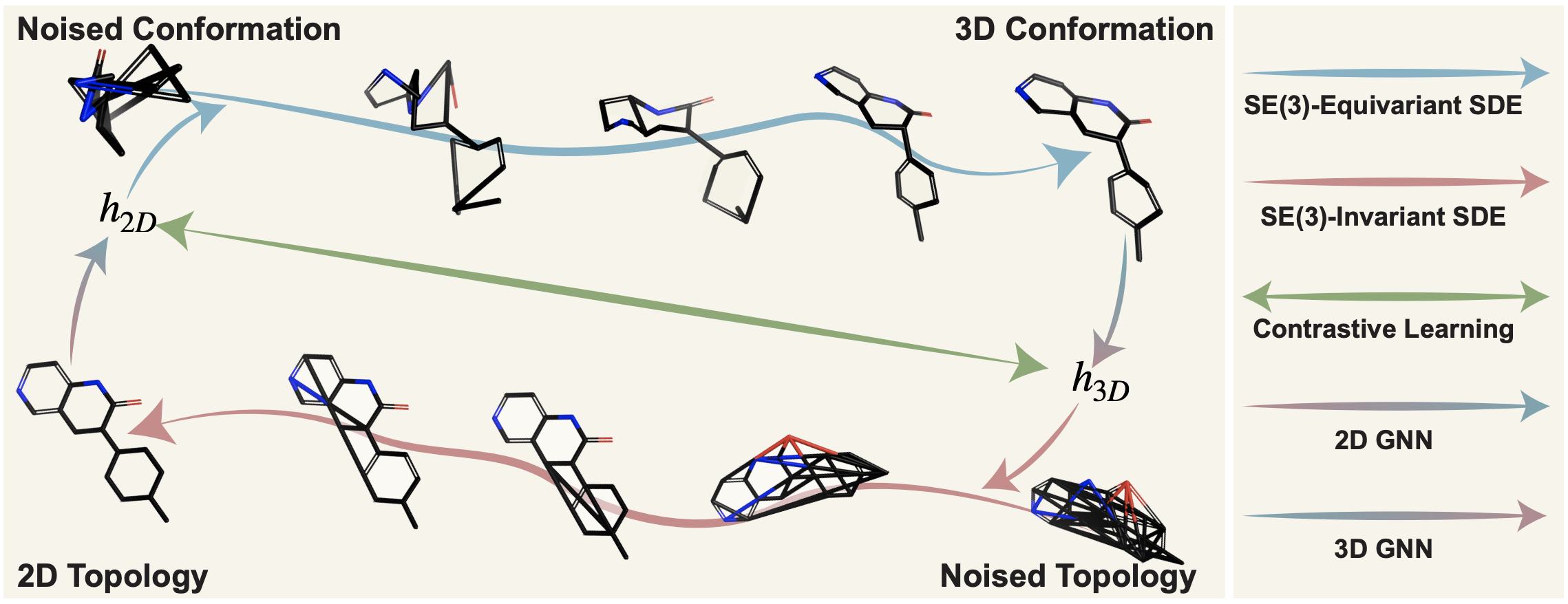}
\vspace{-5ex}
\caption{\small
Illustration of the \model{} pretraining. It is composed of one contrastive learning and two generative learning objectives. Contrastive learning aims to align the 2D topological and 3D conformational representations for the same molecule. The two generative learning objectives are molecule conditional generation from 2D topology to 3D conformation and from 3D conformation to 2D topology, respectively. The generative modeling from topology to conformation is an SE(3)-equivariant diffusion process that satisfies the physical symmetry in molecule geometries. The other direction from conformation to topology is an SE(3)-invariant diffusion process since only the invariant type-0 features (nodes and edges) are considered. We further include \href{https://chao1224.github.io/MoleculeSDE}{a demo}, showing the trajectory of this process.\looseness=-1
}
\label{fig:pretraining_pipeline}
\vspace{-2ex}
\end{figure*}

Molecules can be naturally represented as topological graphs, where atoms and covalent bonds are the nodes and edges. Additionally, molecular structures (a.k.a. \textit{conformations}) can be treated as 3D geometric graphs, where the atoms are the point clouds in the 3D Euclidean space. Based on such two modalities, tremendous representation methods have been proposed in a supervised setting~\cite{gilmer2017neural,thomas2018tensor}. Further, by leveraging a large number of molecule datasets~\cite{irwin2012zinc,hu2020open,axelrod2022geom,xu2021molecule3d}, \textit{molecule pretraining strategies}~\cite{hu2019strategies,sun2022rethinking,liu2022multi} have proven their effectiveness in learning robust and expressive molecule representations. However, most such pertaining works focus on exploring the 2D topology modality, and typical algorithms include reconstructing the masked substructures~\cite{hu2019strategies,liu2019n} and aligning the positive subgraph pairs and contrasting the negative pairs~\cite{sun2019infograph,you2020graph,wang2021molclr} simultaneously. Recently, there have also been successful explorations~\cite{liu2022molecular,jiao20223d} on the 3D geometric pretraining, where the key idea is to reconstruct the masked distances or coordinates through a group symmetric reconstruction operation.

Nevertheless, despite its shown potential for forming high-quality representations, multi-modal pretraining over the molecular 2D topologies and 3D conformations has been under-explored. GraphMVP~\cite{liu2021pre} is the first to build a unified multi-modal self-supervised learning (SSL) paradigm. It introduces contrastive and generative learning to estimate the mutual information (MI) between the two modalities. Contrastive learning treats the topologies and conformations as positive if and only if they are referred to the same molecules. It aims to align the positive pairs while distributing the negative pairs. Such a contrastive idea has also been studied in~\cite{stark20223d}. More encouragingly, GraphMVP demonstrates the benefits of generative SSL, which aims at reconstructing the conformations from the topologies (or vice versa) by introducing a proxy to the evidence lower bound of MI estimation, {\ie}, the \textit{variational representation reconstruction (VRR)}. VRR transforms the reconstruction on the data space ({\ie}, molecular topologies or conformations) to the representation space that compresses input features. Hence, the use of such a proxy can result in the loss of important structural information of molecules. 

\textbf{Our Approach.} To address the aforementioned issue, we propose \model{}, a multi-modal pretraining method on molecules' topologies and conformations. As illustrated in~\Cref{fig:pretraining_pipeline}, \model{} contains both contrastive and generative SSLs. The former adopts the EBM-NCE in~\cite{liu2021pre}, and the latter leverages the stochastic differential equation (SDE) framework~\cite{song2020score,du2022flexible}. Such design brings in two main benefits. First, for the objective function, the generative loss in GraphMVP is a proxy to the MI, while the diffusion process in \model{} leads to a more accurate MI estimation with less information loss. We list a brief performance comparison of two methods in~\Cref{tab:motivation} and will provide theoretical insights that \model{} can lead to a more accurate MI estimation. Second, the SDE-based generative SSL enables prosperous downstream tasks. For example, \model{} enables conformation generation (CG) on tasks where only 2D topologies are available~\cite{wu2018moleculenet}. Based on this, we can apply more advanced geometric modeling methods for prediction. As shown in~\Cref{sec:experiments}, such generated conformations lead to improved predictive performance over existing CG methods~\cite{shi2021learning,du2022se}.

\begin{table}[t!]
\vspace{-1.5ex}
\centering
\small
\caption{
\small
Downstream tasks' performance comparison with \textit{merely} generative pretraining. The complete results are in~\Cref{sec:ablation_studies}.
}
\label{tab:motivation}
\begin{adjustbox}{max width=0.48\textwidth}
\small
\setlength{\tabcolsep}{4pt}
\centering
\begin{tabular}{l l c c c c c c c c c}
\toprule
Model & Tox21 $\uparrow$ & MUV  $\uparrow$ & Bace $\uparrow$ & GAP  $\downarrow$ & U0  $\downarrow$ & Aspirin $\downarrow$\\
\midrule
VRR (GraphMVP) & 73.6 & 75.5 & 72.7 & 44.64 & 13.96 & 1.177 \\
SDE (\model{}) & \textbf{75.6} & \textbf{80.1} & \textbf{79.0} & \textbf{42.75} & \textbf{11.85} & \textbf{1.087} \\
\bottomrule
\end{tabular}
\end{adjustbox}
\vspace{-3ex}
\end{table}

The core components of the proposed \model{} are the two SDE generative processes. The first SDE aims to convert from topology to conformation. This conversion needs to satisfy the physical nature of molecules: the molecules' physical and chemical attributes  need to be equivariant to the rotations and translations in the 3D Euclidean space, {\ie}, the \textbf{SE(3)-equivariant and reflection-antisymmetric} property (\Cref{sec:app:group_symmetry}), and we use \textbf{SE(3)-equivariance} for short.  We note that existing topology to conformation deep generative methods are either SE(3)-invariant~\cite{shi2021learning} or not SE(3)-equivariant~\cite{zhu2022direct}. We propose an SE(3)-equivariant diffusion process by building equivariant local frames. The inputs of local frames are SE(3)-equivariant vector features ({\eg}, the atom coordinates), and the local frames transform them into three SE(3)-invariant features, which will be transformed back to the equivariant data using tensorziation~\cite{du2022se}. The second SDE targets the conformation to topology reconstruction task in the discrete topological space. The main challenge for this task is to adopt the diffusion model for discrete data generation.  We follow the recent work on graph diffusion generation~\cite{jo2022score}, where the joint generation of atom and bond leads to better estimation.

To the best of our knowledge, we are the first to build an \textbf{SE(3)-equivariant and reflection-antisymmetric} SDE for the topology to conformation generation and also the first to devise an SE(3)-invariant SDE for the conformation to topology generation for representation learning. We also note that our proposed \model{} is agnostic to the backbone representation methods since the SDE process is disentangled with the representation function, as illustrated in~\Cref{fig:pretraining_pipeline}.\looseness=-1

Our main contributions include: (1) We propose a group symmetric pretraining method, \model{}, on the 2D and 3D modalities of molecules. (2) We provide theoretical insights on the tighter MI estimation of \model{} over previous works. (3) We show that \model{} enables prosperous downstream tasks. (4) We empirically verify that \model{} retains essential knowledge from both modalities, resulting in state-of-the-art performance on 26 out of 32 downstream tasks compared with 17 competitive baselines.\looseness=-1

\section{Related Work} \label{sec:related}

\textbf{Molecule SSL pretraining on a single modality.} The pretraining on \textit{2D molecular topology} shares common ideas with the general graph pretraining~\cite{sun2022rethinking,wang2022evaluating}. One classical approach~\cite{hu2019strategies,liu2019n} is to mask certain key substructures of molecular graphs and then perform the reconstruction in an auto-encoding manner. Another prevalent molecule pretraining method is contrastive learning~\cite{oord2018representation}, where the goal is to align the views from the positive pairs and contrast the views from the negative pairs simultaneously. For example, ContexPred~\cite{hu2019strategies} constructs views based on different radii of neighborhoods, Deep Graph InfoMax~\cite{velivckovic2018deep} and InfoGraph~\cite{sun2019infograph} treat the local and global graph representations as the two views, MolCLR~\cite{wang2022molecular} and GraphCL~\cite{You2020GraphCL} create different views using discrete graph augmentation methods.

Recent studies start to explore the \textit{3D geometric pretraining} on molecules. GeoSSL~\cite{liu2022molecular} proposes maximizing the mutual information between noised conformations using an SE(3)-invariant denoising score matching, and a parallel work ~\cite{zaidi2022pre} is a special case of GeoSSL using only one denoising layer. 3D-EMGP~\cite{jiao20223d} is also a parallel work, but it is E(3)-equivariant, which needlessly satisfies the reflection-equivariant constraint in molecular conformation distribution.

\textbf{Molecule SSL pretraining on multiple modalities.} The GraphMVP~\cite{liu2021pre} proposes one contrastive objective (EBM-NCE) and one generative objective (variational representation reconstruction, VRR) to optimize the mutual information between the topological and conformational modalities. Specifically for VRR, it is a proxy loss to the evidence lower bound (ELBO) by doing the reconstruction in the representation space, which may risk losing critical information. 3D InfoMax~\cite{stark20223d} is a special case of GraphMVP, where only the contrastive loss is considered. Notice that MoleculeSDE solves the approximation issue in GraphMVP with two SDE processes. 

\section{Preliminaries} \label{sec:preliminaries}

\textbf{2D topological molecular graph.}
A topological molecular graph is denoted as $g_{\twoD{}} = (\mX, \mE)$, where $\mX$ is the atom attribute matrix and $\mE$ is the bond attribute matrix. The 2D graph representation with graph neural network (GNN) is:
\begin{equation} \label{eq:2d_gnn}
\mH_{\twoD{}} = \text{GNN-\twoD{}}(T_{\twoD{}}(g_{\twoD{}})) = \text{GNN-\twoD{}}(T_{\twoD{}}(\mX, \mE)),
\end{equation}
where $T_{\twoD{}}$ is the data transformation on the 2D topology, and GNN-\twoD{} is the representation function. $\mH_{\twoD} = [h_{\twoD{}}^0, h_{\twoD{}}^1, \hdots]$, where $h_{\twoD{}}^i$ is the $i$-th node representation.

\textbf{3D conformational molecular graph.} 
The molecular conformation is denoted as $g_{\threeD{}} = (\mX, \mR)$, where $\mR = \{\vr^1, \vr^2, \hdots\}$ is the collection of 3D coordinates of atoms. The conformational representation is:
\begin{equation} \label{eq:3d_gnn}
\mH_{\threeD{}} = \text{GNN-\threeD{}}(T_{\threeD{}}(g_{\threeD{}})) = \text{GNN-\threeD{}}(T_{\threeD{}}(\mX, \mR)),
\end{equation}
where $T_{\threeD{}}$ is the data transformation on the 3D geometry, and GNN-\threeD{} is the representation function. $\mH_{\threeD{}} = [h_{\threeD{}}^0, h_{\threeD{}}^1, \hdots]$, where $h_{\threeD{}}^i$ is the $i$-th node representation. In our approach, we take the masking as the transformation for both 2D and 3D GNN, and the masking ratio is $M$. In what follows, we use $\vx$ and $\vy$ for the 2D and 3D graphs for notation simplicity, {\ie}, $\vx \triangleq g_\twoD{}$ and $\vy \triangleq g_\threeD{}$.

\textbf{SE(3)-Equivariance and Reflection-Antisymmetry.}
Two 3D geometric graphs $\mR_1$ and $\mR_2$ are SE(3)-isometric if there exists an element $g \in$ SE(3) such that $\mR_2 = g \mR_1$, where $g \in$ SE(3) is a 3D rotation or translation acting on each node (atom) of $\mR_1$. In this article, we will consider vector-valued functions defined on the 3D molecular graph. Specifically, given a conformation-related function $f(\vr): \mathbf{R}^3 \rightarrow  \mathbf{R}^3$ on the graph $\mR$, we say it's equivariant if\looseness=-1
\begin{equation} \label{eq:equivariance_and_reflection_symmetric}
\begin{aligned}
f(g\vr) = g f(\vr) 
\end{aligned}
\end{equation}
for arbitrary $g \in$ SE(3). Since different chiralities can lead to different chemical properties~\cite{clayden2012organic,cotton1991chemical}, the function $f(\vr)$ we consider in this article is SE(3)-equivariant and reflection-antisymmetric.

\textbf{Stochastic Differential Equation (SDE).} The score-based generative modeling with stochastic differential equations (SDEs)~\cite{song2020score} provides a novel and expressive tool for distribution estimation. It is also a united framework including denoising score matching~\cite{vincent2011connection,song2019generative} and denoising diffusion~\cite{ho2020denoising}. In general, these methods can be split into two processes: the forward and backward processes. The forward process is a parameter-free deterministic process, and it diffuses a data point $\vx$ into a random noise by adding noises, as
\begin{equation} \label{eq:SDE_forward}
d\vx_t = f(\vx_t, t) dt + g(t) dw_t,
\end{equation}
with $f(\vx_t,t)$ the vector-value drift coefficient, $g(t)$ the diffusion coefficient, and $w_t$ the Wiener process. Note that \cref{eq:SDE_forward} induces a family of densities $\vx_t \sim p_t(\cdot)$. On the other hand, the backward process generates real data from the stationary distribution of~\Cref{eq:SDE_forward} by evolving along the following SDE:
\begin{equation} \label{eq:SDE_backward}
d\vx_t = [f(\vx_t,t) - g(t)^2 \nabla_\vx \log p_t(\vx_t)] dt + g(t) dw_t .
\end{equation}
Thus, the learning objective is to estimate $\nabla_\vx \log p_t(\vx)$. This derivative term is called \textit{score} in the literature of score matching~\cite{vincent2011connection,song2019generative,song2020score}. We will use this framework in \model{} to generate 3D conformation and 2D topology.

\section{The \model{} Method} \label{sec:method}
In~\Cref{sec:method_overview}, we first provide the mutual information (MI) aspect in molecule multi-modal pretraining, and then we present the limitation of VRR. We discuss two SDE models as generative pretraining in~\Cref{sec:generative_ssl_2D_to_3D,sec:generative_ssl_3D_to_2D}, respectively. The ultimate learning objective and inference process are illustrated in~\Cref{sec:ultimate_objective}. Additionally in~\Cref{sec:theoretical_analysis}, we provide theoretical insights on how \model{} obtains a more accurate MI estimation.

\subsection{An Overview from Mutual Information Perspective} \label{sec:method_overview}
Mutual information (MI) measures the non-linear dependency between random variables, and it has been widely adopted as the principle for self-supervised pretraining~\cite{oord2018representation,hjelm2018learning}. The expectation is that, by maximizing the MI between modalities, the learned representation can keep the most shared information. Thus, MI-guided SSL serves as an intuitive and powerful framework for representation pretraining.

GraphMVP~\cite{liu2021pre} transforms the MI maximization into the summation of two conditional log-likelihoods:
\begin{equation} \label{eq:MI_objective}
\mathcal{L}_{\text{MI}} = \frac{1}{2} \mathbb{E}_{p(\vx,\vy)} \big[ \log p(\vy|\vx) + \log p(\vx|\vy) \big].
\end{equation}
GraphMVP~\cite{liu2021pre} further solves~\Cref{eq:MI_objective} by proposing a contrastive loss (EBM-NCE) and a generative loss (variational representation reconstruction, VRR). VRR conducts the reconstruction on the representation space, {\ie}, from 2D (3D) data to the 3D (2D) representation space (details in~\Cref{sec:app:MI_conditional_likelihoods}). The main advantage of VRR is its simple implementation without topology or conformation reconstruction, yet the trade-off is that it can lose critical information since it is only a proxy solution to generative learning.\looseness=-1

Thus, to this end, we raise one question: \textit{If there exists a more accurate conditional density estimation method for generative pretraining?} The answer is yes, and we propose \model{}. \model{} utilizes two stochastic differential (SDE) models to estimate~\Cref{eq:MI_objective}. SDE is a broad generative model class~\cite{karras2022elucidating} where a neural network is used to model the score~\cite{song2021train} of various levels of noise in a diffusion process~\cite{ho2020denoising}. To adapt it in \model{}, we propose an SDE from 2D topology to 3D conformation (\Cref{sec:generative_ssl_2D_to_3D}) and an SDE from 3D conformation to 2D topology (\Cref{sec:generative_ssl_3D_to_2D}). We also want to highlight that such reconstructions are challenging, as both the 2D topologies and 3D conformations are highly structured: the 2D topologies are permutation invariant, and the 3D conformations additionally obey the SE(3)-equivariance.

\model{} has three main advantages. (1) \model{} is a powerful generative pretraining method, as the SDE models have shown promising performance in applications including image generation~\cite{ho2020denoising,song2020score} and geometric representation~\cite{liu2022molecular,zaidi2022pre}. (2) \model{} is a more accurate estimation to~\Cref{eq:MI_objective} than previous methods. Thus, the pretrained representation contains more critical information with a more accurate MI estimation. (3) \model{} enables prosperous downstream tasks, like topology to conformation generation for property prediction (\Cref{sec:downstream_2Dto3D}).

\subsection{An SE(3)-Equivariant Conformation Generation} \label{sec:generative_ssl_2D_to_3D}
The first objective we consider is the conditional generation from topology to conformation, $p(\vy|\vx)$. One thing to highlight is that the molecule 3D conformation needs to satisfy the physical property, {\ie}, it needs to be equivariant to the rotation and transition in the 3D Euclidean space, which is known as the \textit{SE(3)-equivariance and reflection-antisymmetry} property (\Cref{sec:app:group_symmetry}). Notice that for notation simplicity, we may call it \textit{SE(3)-equivariance}, and only expand into details in the "Local frame" paragraph below.

The core module in SDE is the score network, $S_\theta^{\twoD{} \rightarrow \threeD{}}$. To satisfy the physical nature of molecule 3D structure, such a score network needs to be SE(3)-equivariant. Specifically, the input includes the 2D graph $\vx$, the noised 3D information $\vy_t$ at time $t$, and the time $t$. The output is the SE(3)-equivariant 3D scores at time $t$, accordingly. The goal is to use $S_\theta^{\twoD{} \rightarrow \threeD{}}$ to estimate the score $\nabla \log p_t (\vy_t | \vx)$.

To learn $ p(\vy|\vx)$, we formulate it as solving an SDE problem. Then based on the score network, the training objective is:
\begin{equation}
\begin{aligned}
\mathcal{L}^{\twoD{} \rightarrow \threeD{}} 
& = \mathbb{E}_{\vx,\vy} \mathbb{E}_t \mathbb{E}_{\vy_t | \vy} \\
& \Big[ \Big\| \nabla_{\vy_t}\log p_t (\vy_t| \vy, \vx)- S_\theta^{\twoD{} \rightarrow \threeD{}} (\vx , \vy_t, t) \Big\|_2^2 \Big].
\end{aligned}
\end{equation}

\textbf{Coordinate reconstruction.} Notice that both 2D topologies and 3D conformations share the same atom information (atom types), so in this subsection specifically, by reconstructing $\vy$, we are referring to reconstructing the coordinates $\mR$. Thus, the objective function becomes:
\begin{equation}
\begin{aligned}
\mathcal{L}^{\twoD{} \rightarrow \threeD{}}
& = \mathbb{E}_{\vx,\mR} \mathbb{E}_t \mathbb{E}_{\mR_t | \mR} \\
& \Big[ \Big\| \nabla_{\mR_t}\log p_t (\mR_t| \mR, \vx)- S_\theta^{\twoD{} \rightarrow \threeD{}} (\vx, \mR_t, t) \Big\|_2^2 \Big],
\end{aligned}
\end{equation}

\textbf{Local frame.} Before going into details of the score network, we want to introduce the SE(3)-equivariant \& reflection-antisymmetric local frame (\Cref{sec:app:group_symmetry}). Such equivariant frames are introduced to fill in the gap between invariant 2D features and the output SE(3) vector field. It is equivalent to a 3D coordinate system that transforms equivariantly with the geometric graph. Through equivariant frames, we can project noised 3D coordinates into invariant scalars (isomers are projected differently), such that they are ready to be combined with invariant 2D features. On the other hand, by projecting back the 2D graph's invariant predictions into an equivariant frame, our final output can transform equivariantly with respect to global rotation and translation. We leave the precise formulations of our local frames in~\Cref{sec:app:group_symmetry}. Briefly in \model{}, we focus on the equivariant frame attached on each edge $(\vr^i,\vr^j)$:
\begin{equation}
\begin{aligned}
\vt_{\localframe{}}^{ij} = \text{Local-Frame}(\vr^i, \vr^j).\\
\end{aligned}
\end{equation}

\textbf{SE(3)-equivariant score network.} Then we introduce how to build an SE(3)-equivariant score network based on the local frame. We first concat the atom representations $h_{\twoD{}}$ into the atom pairwise representations, as $e^{ij}_{\twoD{}} = \text{MLP} (\text{concat}\{h_{\twoD{}}^i \ ||\  h_{\twoD{}}^j\})$ for the $i$-th and $j$-th atoms. Then the 2D pairwise representations are further added to the 3D pairwise representations $e^{ij}_{\threeD{}} = \text{projection}_{\vt_{\localframe{}}^{ij}} (\vr^i, \vr^j)$, produced by the equivariant frames. Based on such frame feature, the final \textbf{invariant} edge feature $e^{ij}$ is defined by:
\begin{equation}
e^{ij} = \text{rbf}(r^{ij}) \odot e^{ij}_{\twoD{}} + e^{ij}_{\threeD{}},
\end{equation}
where $r^{ij}$ denotes the relative distance between the $i$-th and $j$-th atoms, and we use the radial basis function (RBF) to embed such distance features. Note that the input 3D coordinates and the corresponding distance matrix $\{r^{ij}\}$ are based on the diffused positions at a given diffusion step, rather than the ground truth 3D conformation.  

Then we process $e^{ij}$ through multiple graph attention layers~\cite{shi2020masked}: $h^{ij} = \text{Attention} (e^{ij})$. Finally, by pairing the invariant aggregated edge features $h^{ij}$ with our SE(3)-equivariant frames $\vt_{\localframe{}}^{ij}$, we get the vector-valued score function: $S(\vr^i) = \sum_j h^{ij} \odot \vt_{\localframe{}}^{ij}$. Here, our equivariant construction guarantees that the output vector field is SE(3)-equivariant and reflection-antisymmetric.

\textbf{Discussions.} We want to clarify the following points between molecule geometric modeling ($h_\threeD{}$) and score network ($S_\theta^{\twoD{} \rightarrow \threeD{}}$). (1) The score network proposed here is SE(3)-equivariant and reflection-antisymmetric. The input to the score network is the topology and diffused conformation, so it cannot be shared with the molecule geometric modeling, where the input is the ground-truth 3D conformation. (2) To the best of our knowledge, we are the first to propose the SE(3)-equivariant and reflection-antisymmetric SDE for the topology to conformation generation task.

\subsection{An SE(3)-Invariant Topology Generation} \label{sec:generative_ssl_3D_to_2D}
The second objective is to reconstruct the 2D topology from 3D conformation, {\ie}, $p(\vx|\vy)$. Note that the 2D topology information (atoms and bonds) belongs to the type-0 feature~\cite{thomas2018tensor}, thus such a generative process should satisfy the SE(3)-invariance property. If we formulate it as an SDE problem, then the training objective is:
\begin{equation}
\begin{aligned}
\mathcal{L}^{\threeD{} \rightarrow \twoD{}} 
& = \mathbb{E}_{\vy, \vx} \mathbb{E}_t \mathbb{E}_{\vx_t | \vx}\\
& \Big[ \Big\| \nabla_{\vx_t}\log p_t (\vx_t | \vx, \vy) - S_\theta^{\threeD{} \rightarrow \twoD{}} (\vy, \vx_t, t) \Big\|_2^2 \Big],
\end{aligned}
\end{equation}
where $S_\theta^{\threeD{} \rightarrow \twoD{}}$ is the score network.

\textbf{SE(3)-invariant score network.}
For modeling $S_\theta^{\threeD{} \rightarrow \twoD{}}$, it needs to satisfy the SE(3)-invariance symmetry property. The inputs are 3D conformational representation $\vy$, the noised 2D information $\vx_t$ at time $t$, and time $t$. The output of $S_\theta^{\threeD{} \rightarrow \twoD{}}$ is the invariant 2D score function at time $t$, as $\nabla \log p_t (\vx_t | \vy)$. As introduced in~\Cref{sec:preliminaries}, the diffused 2D information contains two parts: $\vx_t = (\mX_t, \mE_t)$, so the corresponding forward SDE is a joint variant of \Cref{eq:SDE_forward}:
\begin{align}
\begin{cases}
    \! \mathrm{d}\mX_t \! = \! f_{1,t}(\mX_t,\mE_t) dt + g_{1}(t)\mathrm{d}w^1_t, \\[5pt]
    \! \mathrm{d}\mE_t = \! f_{2,t}(\mX_t,\mE_t) dt + g_{2}(t)\mathrm{d}w^2_t,
\end{cases}
\end{align}
where $w^1_t$ and $w^2_t$ are two independent Brownian motion. Then the score network $S_\theta^{\threeD{} \rightarrow \twoD{}}$ is also decomposed into two parts for the atoms and bonds: $S_\theta^{\mX_t}(\vx_t)$ and $S_\theta^{\mE_t}(\vx_t)$.

Similar to the topology to conformation generation procedure, we first merge the 3D representation $\vz_y$ with the diffused atom feature ${\mX_t}$ as $H_0 = \text{MLP}(\mX_t) + \mH_{\vy}$. Then we apply a GCN as the score network to estimate the node-level score, as $S_\theta^{\mX_t}(\vx_t) =  \text{MLP} (\text{concat}\{H_0 || \cdots || H_L\})$, where $H_{i+1} = \text{GCN}(H_i,\mE_t)$ and $L$ is the number of GCN layer. On the other hand, the edge-level score is modeled by an unnormalized dot product attention $S_\theta^{\mE_t}(\vx_t) = \text{MLP} (\{\text{Attention}(H_i)\}_{0 \leq i \leq L})$.

\subsection{Learning and Inference of \model{}} \label{sec:ultimate_objective}
\textbf{Learning.} In addition to the two generative objectives, we also consider a contrastive loss, EBM-NCE~\cite{liu2021pre}. EBM-NCE can be viewed as another way to approximate the mutual information $I(X;Y)$, and it is expected to be complementary to the generative SSL. Therefore, our final objective is
\begin{equation}
\mathcal{L}_{\text{\model{}}} = \alpha_1 \mathcal{L}_{\text{Contrastive}} + \alpha_2 \mathcal{L}^{\twoD{} \rightarrow \threeD{}} + \alpha_3 \mathcal{L}^{\threeD{} \rightarrow \twoD{}},   
\end{equation}
where $\alpha_1, \alpha_2, \alpha_3$ are three coefficient hyperparameters.

\textbf{Inference.} After we train the SDE model from 2D topologies to 3D conformations, we can generate 3D conformations from a 2D topology by `reversing' the forward SDE. More precisely, we take the Predictor-Corrector sampling method~\cite{song2020score} as tailored to our continuous framework. Further, generating 3D conformations from 2D topologies enable us to conduct prosperous downstream tasks such as property prediction with 2D and 3D data.

\subsection{Theoretical Insights of \model{}} \label{sec:theoretical_analysis}
Since the two terms in~\Cref{eq:MI_objective} are in the mirroring direction, here we take $\vx|\vy$ for theoretical illustrations. The other direction can be obtained similarly. We adapt the continuous diffusion framework proposed in~\cite{song2020score}, in which the Markov chain denoising generative model (DDPM) is included as a discretization of the continuous Ornstein–Uhlenbeck process. The diffusion framework originates from the noised score-matching scheme~\cite{vincent2011connection} of training the energy-based model (EBM),  in which the authors introduced a noised version of $p(\vy)$ by adding noise to each data point $\Tilde{\vy} = \vy + \epsilon$, where $\epsilon$ is sampled from a scaled normal distribution. Then,
{
\begin{align}
\label{one step}
&\min_{\theta}\mathbb{E}_{p(\Tilde{\vy})}\norm{\nabla_{\vy} \log p(\Tilde{\vy}|\vx) - \nabla_{\vy} \log p_{\theta} (\Tilde{\vy}| \vx)}_2^2 \\ \nonumber
= &\min_{\theta}\mathbb{E}_{p(\Tilde{\vy},\vy)}\norm{\nabla_{\vy} \log p(\Tilde{\vy}|\vy,\vx) - \nabla_{\vy} \log p_{\theta} (\Tilde{\vy}|\vx)}_2^2+ C, 
\end{align} 
}
where the conditional score function $\nabla_{\vy}\log p(\Tilde{\vy}|\vy,\vx)$ is analytically tractable. The diffusion generative model further pushes the one-step (\ref{one step}) to a continuous noising process from raw data $\vy$ to $\vy_t$ for $0 \leq t \leq T$. We call $\vy_t$ the nosing (forward) diffusion process starting at $y$, which is usually formulated as the solution of a stochastic differential equation. Then, the corresponding (continuous) score matching loss is:
{
\fontsize{7.5}{1}\selectfont
\begin{equation} \label{multi step} 
\min_{\theta} \mathbb{E}_t \mathbb{E}_{\vy} \mathbb{E}_{\vy_t | \vy}\norm{\nabla_{\vy_t}\log p_t (\vy_t|\vy,\vx)-\nabla_{\vy_t}\log p_{\theta,t} (\vy_t|\vx)}_2^2.
\end{equation}
}
\normalsize

It's worth mentioning that the weighted continuous score matching is equivalent to learning the infinitesimal reverse of the noising process from $t$ to $t + \Delta t$ for each time $t$, which greatly reduces the difficulty of recovering $p_{data}$ from the white noise in one shot~\cite{ho2020denoising}.

To make a connection between \Cref{multi step} and  \Cref{eq:MI_objective}, it's crucial to relate score matching with \textbf{maximal log-likelihood} method. To solve this problem, \citep{NEURIPS2021_c11abfd2} defined a key quantity (ELBO) $\mathcal{E}^{\infty}_{\theta}(\vy|\vx)$ as a functional on the infinite-dimensional path space (consists of all stochastic paths starting at $\vy$). Then, the authors show that 
\begin{equation}
    \mathbb{E}_{p(\vy)}\log p_{\theta,T} (\vy|\vx) \ge \mathbb{E}_{p(\vy)}\mathcal{E}^{\infty}_{\theta}(\vy|\vx),
\end{equation}
where the probability $p_{\theta,T}(\vy|\vx)$ corresponds to the marginal distribution of a parameterized (denoted by $\theta$) SDE at time $T$. Moreover, the ELBO $\mathbb{E}_{p(\vy)}\mathcal{E}^{\infty}_{\theta}(\vy|\vx)$ is equivalent to the score matching loss. Therefore, training the diffusion model is equivalent to maximizing a lower bound of the likelihood defined by SDEs. Since the variational capacity of the infinite-dimensional SDE space is larger than previous models, we expect to find a better estimation of \cref{eq:MI_objective}. 

\section{Experiments} \label{sec:experiments}

\model{} enables both a pretrained 2D and 3D representation and can be further fine-tuned toward the downstream tasks.
Meanwhile, another main advantage of \model{} is that it also learns an SDE model from topology to conformation. Such a design enables us to adopt more versatile downstream tasks. For instance, there is a wide range of molecular property prediction tasks~\cite{wu2018moleculenet} considering only the 2D topology, yet the 3D conformation has proven to be beneficial towards such property prediction tasks~\cite{liu2021pre}. Thus, with the pretrained generative model $p(\vy|\vx)$, we can generate the corresponding 3D structure for each molecule topology and apply the pretrained 3D encoders, which is expected to improve the performance further. A visual illustration of such three categories of downstream tasks is in~\Cref{fig:downstream_pipeline}.

\begin{table*}[ht!]
\caption{
Results for molecular property prediction tasks (with 2D topology only). 
For each downstream task, we report the mean (and standard deviation) ROC-AUC of 3 seeds with scaffold splitting.
The best and second best results are marked \textbf{\underline{bold}} and \textbf{bold}, respectively.
}
\label{tab:main_results_moleculenet_2D}
\centering
\begin{adjustbox}{max width=\textwidth}
\small
\setlength{\tabcolsep}{4pt}
\centering
\begin{tabular}{l c c c c c c c c c}
\toprule
Pre-training & BBBP $\uparrow$ & Tox21 $\uparrow$ & ToxCast $\uparrow$ & Sider $\uparrow$ & ClinTox $\uparrow$ & MUV $\uparrow$ & HIV $\uparrow$ & Bace $\uparrow$ & Avg $\uparrow$ \\
\midrule
-- (random init) & 68.1$\pm$0.59 & 75.3$\pm$0.22 & 62.1$\pm$0.19 & 57.0$\pm$1.33 & 83.7$\pm$2.93 & 74.6$\pm$2.35 & 75.2$\pm$0.70 & 76.7$\pm$2.51 & 71.60 \\
AttrMask & 65.0$\pm$2.36 & 74.8$\pm$0.25 & 62.9$\pm$0.11 & \textbf{61.2$\pm$0.12} & \underline{\textbf{87.7$\pm$1.19}} & 73.4$\pm$2.02 & 76.8$\pm$0.53 & 79.7$\pm$0.33 & 72.68 \\
ContextPred & 65.7$\pm$0.62 & 74.2$\pm$0.06 & 62.5$\pm$0.31 & \underline{\textbf{62.2$\pm$0.59}} & 77.2$\pm$0.88 & 75.3$\pm$1.57 & 77.1$\pm$0.86 & 76.0$\pm$2.08 & 71.28 \\
InfoGraph & 67.5$\pm$0.11 & 73.2$\pm$0.43 & 63.7$\pm$0.50 & 59.9$\pm$0.30 & 76.5$\pm$1.07 & 74.1$\pm$0.74 & 75.1$\pm$0.99 & 77.8$\pm$0.88 & 70.96 \\
MolCLR & 66.6$\pm$1.89 & 73.0$\pm$0.16 & 62.9$\pm$0.38 & 57.5$\pm$1.77 & 86.1$\pm$0.95 & 72.5$\pm$2.38 & 76.2$\pm$1.51 & 71.5$\pm$3.17 & 70.79 \\
3D InfoMax & 68.3$\pm$1.12 & 76.1$\pm$0.18 & 64.8$\pm$0.25 & 60.6$\pm$0.78 & 79.9$\pm$3.49 & 74.4$\pm$2.45 & 75.9$\pm$0.59 & 79.7$\pm$1.54 & 72.47 \\
GraphMVP & 69.4$\pm$0.21 & 76.2$\pm$0.38 & 64.5$\pm$0.20 & 60.5$\pm$0.25 & 86.5$\pm$1.70 & 76.2$\pm$2.28 & 76.2$\pm$0.81 & \textbf{79.8$\pm$0.74} & 73.66 \\
\midrule

\model{} (VE) & \underline{\textbf{73.2$\pm$0.48}} & \textbf{76.5$\pm$0.33} & \underline{\textbf{65.2$\pm$0.31}} & 59.6$\pm$0.82 & 86.6$\pm$3.73 & \textbf{79.9$\pm$0.19} & \textbf{78.5$\pm$0.28} & \underline{\textbf{80.4$\pm$0.92}} & \textbf{74.98} \\

\model{} (VP) & \textbf{71.8$\pm$0.76} & \underline{\textbf{76.8$\pm$0.34}} & \textbf{65.0$\pm$0.26} & 60.8$\pm$0.39 & \textbf{87.0$\pm$0.53} & \underline{\textbf{80.9$\pm$0.37}} & \underline{\textbf{78.8$\pm$0.92}} & 79.5$\pm$2.17 & \underline{\textbf{75.07}} \\
\bottomrule
\end{tabular}
\end{adjustbox}
\vspace{-2ex}
\end{table*}

\begin{table*}[htb]
\setlength{\tabcolsep}{5pt}
\fontsize{9}{9}\selectfont
\caption{
\small
Results on 12 quantum mechanics prediction tasks from QM9. We take 110K for training, 10K for validation, and 11K for testing. The evaluation is mean absolute error, and the best and the second best results are marked in \underline{\textbf{bold}} and \textbf{bold}, respectively.
}
\label{tab:main_QM9_result}
\begin{adjustbox}{max width=\textwidth}
\begin{tabular}{l c c c c c c c c c c c c}
\toprule
Pretraining & $\alpha$ $\downarrow$ & $\nabla \mathcal{E}$ $\downarrow$ & $\mathcal{E}_\text{HOMO}$ $\downarrow$ & $\mathcal{E}_\text{LUMO}$ $\downarrow$ & $\mu$ $\downarrow$ & $C_v$ $\downarrow$ & $G$ $\downarrow$ & $H$ $\downarrow$ & $R^2$ $\downarrow$ & $U$ $\downarrow$ & $U_0$ $\downarrow$ & ZPVE $\downarrow$\\
\midrule

-- (random init) & 0.060 & 44.13 & 27.64 & 22.55 & 0.028 & 0.031 & 14.19 & 14.05 & 0.133 & 13.93 & 13.27 & 1.749\\


Type Prediction & 0.073 & 45.38 & 28.76 & 24.83 & 0.036 & 0.032 & 16.66 & 16.28 & 0.275 & 15.56 & 14.66 & 2.094\\

Distance Prediction & 0.065 & 45.87 & 27.61 & 23.34 & 0.031 & 0.033 & 14.83 & 15.81 & 0.248 & 15.07 & 15.01 & 1.837\\

Angle Prediction & 0.066 & 48.45 & 29.02 & 24.40 & 0.034 & 0.031 & 14.13 & 13.77 & 0.214 & 13.50 & 13.47 & 1.861\\

3D InfoGraph & 0.062 & 45.96 & 29.29 & 24.60 & 0.028 & 0.030 & 13.93 & 13.97 & 0.133 & 13.55 & 13.47 & 1.644\\

RR & 0.060 & 43.71 & 27.71 & 22.84 & 0.028 & 0.031 & 14.54 & 13.70 & \textbf{0.122} & 13.81 & 13.75 & 1.694\\

InfoNCE & 0.061 & 44.38 & 27.67 & 22.85 & \textbf{0.027} & 0.030 & 13.38 & 13.36 & \underline{\textbf{0.116}} & 13.05 & 13.00 & 1.643\\

EBM-NCE & 0.057 & 43.75 & 27.05 & 22.75 & 0.028 & 0.030 & 12.87 & 12.65 & 0.123 & 13.44 & 12.64 & 1.652\\

3D InfoMax & 0.057 & 42.09 & 25.90 & 21.60 & 0.028 & 0.030 & 13.73 & 13.62 & 0.141 & 13.81 & 13.30 & 1.670\\

GraphMVP & 0.056 & 41.99 & 25.75 & \textbf{21.58} & \textbf{0.027} & \textbf{0.029} & 13.43 & 13.31 & 0.136 & 13.03 & 13.07 & 1.609\\

GeoSSL-1L & 0.058 & 42.64 & 26.32 & 21.87 & 0.028 & 0.030 & 12.61 & 12.81 & 0.173 & 12.45 & 12.12 & 1.696\\

GeoSSL & 0.056 & 42.29 & \underline{\textbf{25.61}} & 21.88 & \textbf{0.027} & \textbf{0.029} & \textbf{11.54} & \textbf{11.14} & 0.168 & \textbf{11.06} & \underline{\textbf{10.96}} & 1.660\\

\midrule

\model{} (VE) & \textbf{0.056} & \textbf{41.84} & 25.79 & 21.63 & \textbf{0.027} & \textbf{0.029} & \underline{\textbf{11.47}} & \underline{\textbf{10.71}} & 0.233 & \underline{\textbf{11.04}} & \textbf{10.95} & \underline{\textbf{1.474}}\\

\model{} (VP) & \underline{\textbf{0.054}} & \underline{\textbf{41.77}} & \textbf{25.74} & \underline{\textbf{21.41}} & \underline{\textbf{0.026}} & \underline{\textbf{0.028}} & 13.07 & 12.05 & 0.151 & 12.54 & 12.04 & \textbf{1.587}\\
\bottomrule
\end{tabular}
\end{adjustbox}
\vspace{-2ex}
\end{table*}

\begin{table*}[tb]
\setlength{\tabcolsep}{5pt}
\fontsize{9}{9}\selectfont
\centering
\caption{
\small
Results on eight force prediction tasks from MD17. We take 1K for training, 1K for validation, and 48K to 991K molecules for the test concerning different tasks. The evaluation is mean absolute error, and the best results are marked in \underline{\textbf{bold}} and \textbf{bold}, respectively.
}
\label{tab:main_MD17_result}
\begin{adjustbox}{max width=\textwidth}
\begin{tabular}{l c c c c c c c c}
\toprule
Pretraining & Aspirin $\downarrow$ & Benzene $\downarrow$ & Ethanol $\downarrow$ & Malonaldehyde $\downarrow$ & Naphthalene $\downarrow$ & Salicylic $\downarrow$ & Toluene $\downarrow$ & Uracil $\downarrow$ \\
\midrule

-- (random init) & 1.203 & 0.380 & 0.386 & 0.794 & 0.587 & 0.826 & 0.568 & 0.773\\


Type Prediction & 1.383 & 0.402 & 0.450 & 0.879 & 0.622 & 1.028 & 0.662 & 0.840\\

Distance Prediction & 1.427 & 0.396 & 0.434 & 0.818 & 0.793 & 0.952 & 0.509 & 1.567\\

Angle Prediction & 1.542 & 0.447 & 0.669 & 1.022 & 0.680 & 1.032 & 0.623 & 0.768\\

3D InfoGraph & 1.610 & 0.415 & 0.560 & 0.900 & 0.788 & 1.278 & 0.768 & 1.110\\

RR & 1.215 & 0.393 & 0.514 & 1.092 & 0.596 & 0.847 & 0.570 & 0.711\\

InfoNCE & 1.132 & 0.395 & 0.466 & 0.888 & 0.542 & 0.831 & 0.554 & 0.664\\

EBM-NCE & 1.251 & 0.373 & 0.457 & 0.829 & 0.512 & 0.990 & 0.560 & 0.742\\

3D InfoMax & 1.142 & 0.388 & 0.469 & 0.731 & 0.785 & 0.798 & 0.516 & 0.640\\

GraphMVP & 1.126 & 0.377 & 0.430 & 0.726 & 0.498 & 0.740 & 0.508 & 0.620\\

GeoSSL-1L & 1.364 & 0.391 & 0.432 & 0.830 & 0.599 & 0.817 & 0.628 & 0.607\\

GeoSSL & \underline{\textbf{1.107}} & 0.360 & 0.357 & 0.737 & 0.568 & 0.902 & \textbf{0.484} & 0.502\\

\midrule

\model{} (VE) & \textbf{1.112} & \underline{\textbf{0.304}} & \underline{\textbf{0.282}} & \textbf{0.520} & \textbf{0.455} & \textbf{0.725} & 0.515 & \underline{\textbf{0.447}}\\

\model{} (VP) & 1.244 & \textbf{0.315} & \textbf{0.338} & \underline{\textbf{0.488}} & \underline{\textbf{0.432}} & \underline{\textbf{0.712}} & \underline{\textbf{0.478}} & \textbf{0.468}\\
\bottomrule
\end{tabular}
\end{adjustbox}
\vspace{-3ex}
\end{table*}

\subsection{Pretraining and Baselines}
\textbf{Dataset.}
For pretraining, we use PCQM4Mv2~\cite{hu2020ogb}. It's a sub-dataset of PubChemQC~\cite{nakata2017pubchemqc} with 3.4 million molecules with both the topological graph and geometric conformations. We are aware of the Molecule3D~\cite{xu2021moleculed} dataset, which is also extracted from PubChemQC~\cite{nakata2017pubchemqc}. Yet, after confirming with the authors, certain mismatches exist between the 2D topologies and 3D conformations. Thus, in this work, we use PCQM4Mv2 for pretraining.

\textbf{Baselines for 2D topology pretraining.}
Enormous 2D topological pretraining methods have been proposed~\cite{liu2021graph,xie2021self,wu2021self,liu2021self}. Recent works~\cite{sun2022rethinking} re-explore the effects of these pretraining methods, and we pick up the most promising ones as follows. AttrMask~\citep{hu2019strategies,liu2019n}, ContexPred~\citep{hu2019strategies}, Deep Graph Infomax~\citep{velivckovic2018deep} and InfoGraph~\cite{sun2019infograph}, MolCLR~\cite{wang2022molecular} and GraphCL~\citep{You2020GraphCL}. The detailed explanations are in~\Cref{sec:related}.

\textbf{Baselines for 3D conformation pretraining.}
The 3D conformation SSL pretraining has been less explored. We adopt the comprehensive baselines from~\cite{liu2022molecular}.
The type prediction, distance prediction, and angle prediction predict the masked atom type, pairwise distance, and triplet angle, respectively.
The 3D InfoGraph predicts whether the node- and graph-level 3D representation are for the same molecule. RR, InfoNCE, and EBM-NCE are to maximize the MI between the conformation and augmented conformation using different objective functions, respectively. GeoSSL-DDM optimizes the same objective function using denoising score matching. Another work~\cite{zaidi2022pre} is a special case of GeoSSL-DDM with one layer of denoising, and we name it GeoSSL-DDM-1L.

\textbf{Baselines for 2D-3D multi-modality pretraining.}
There are two baselines on the 2D-3D multi-modal pretraining: vanilla GraphMVP~\cite{liu2021pre} utilizes both the contrastive and generative SSL, and 3D InfoMax~\cite{stark20223d} only uses the contrastive learning part in GraphMVP.

\textbf{Backbone models and \model{}.}
For all the baselines and \model{}, we use the same backbone models to better verify the effectiveness of the pretraining algorithms. We take the GIN model~\cite{xu2018powerful} and SchNet model~\cite{schutt2018schnet} for modeling 2D topology and 3D conformation, respectively. For \model{} training, we consider both the Variance Exploding (VE) and Variance Preserving (VP)  (details in~\Cref{sec:app:SDE}).

\subsection{Downstream with 2D Topology}
We consider eight binary classification tasks from MoleculeNet~\cite{wu2018moleculenet}.
The results are in~\Cref{tab:main_results_moleculenet_2D}. We can observe that \model{} works best on 6 out of 8 tasks, and both the VE and VP version of \model{} pretraining can reach the best average performance.

\subsection{Downstream with 3D Conformation}
We consider 12 tasks from QM9~\cite{ramakrishnan2014quantum} and 8 tasks from MD17~\cite{chmiela2017machine}. QM9 is a dataset of 134K molecules consisting of 9 heavy atoms, and the 12 tasks are related to the quantum properties, such as energies at various settings. MD17 is a dataset on molecular dynamics simulation, and the 8 tasks correspond to 8 organic molecules. The goal is to predict the forces at different 3D positions.
The results are in~\Cref{tab:main_QM9_result,tab:main_MD17_result}, and \model{} can reach the best performance on 9 tasks in QM9 and 7 tasks in MD17.

\begin{figure}[htb!]
\centering
\includegraphics[width=\linewidth]{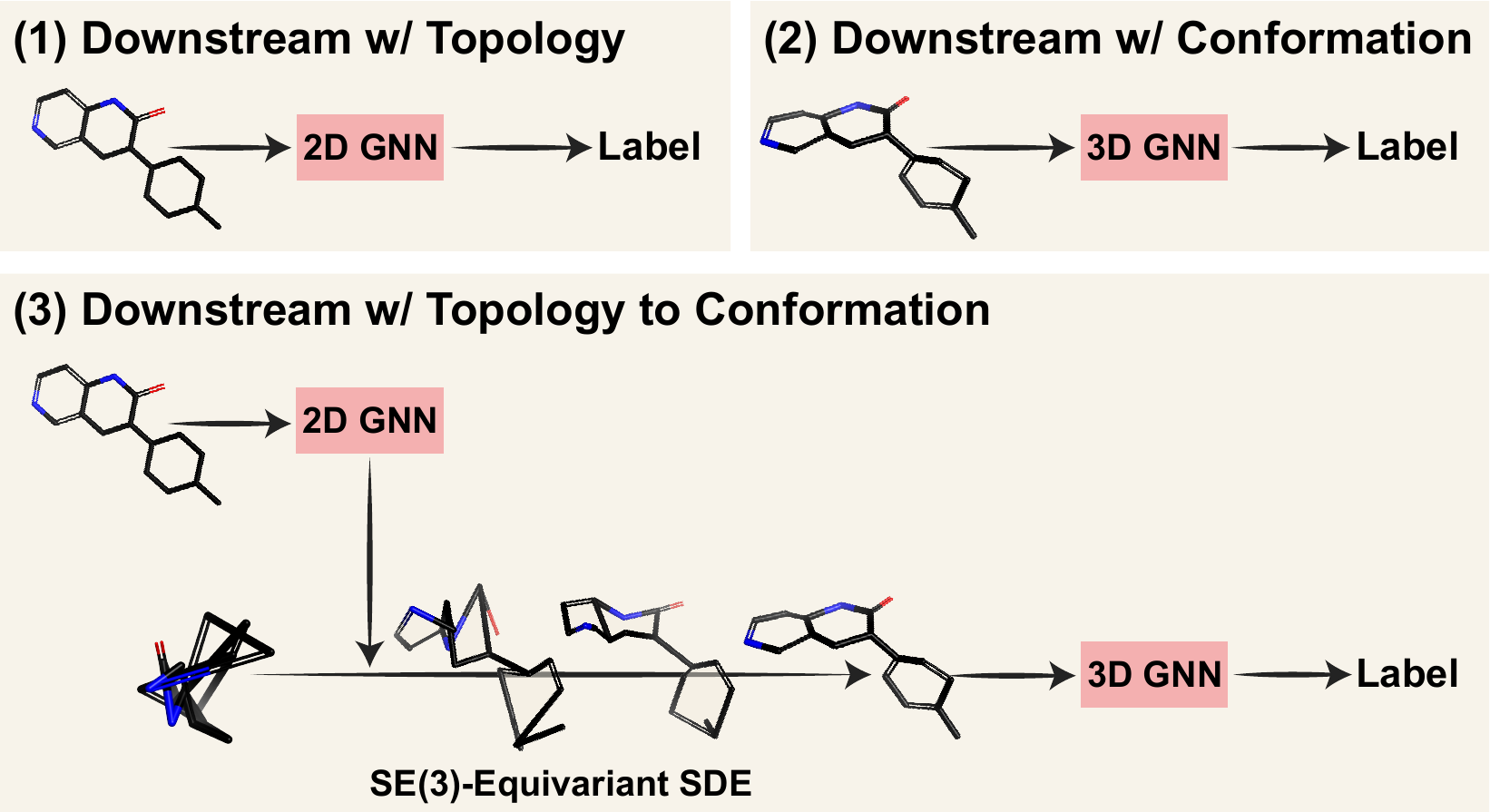}
\vspace{-5ex}
\caption{\small
Illustration on three downstream tasks. The first two cover single-modal information only, and we fine-tune the pretrained 2D and 3D GNN from \model{}, respectively. The last downstream tasks contain topology only, then we use the pretrained 2D GNN and SE(3)-equivariant SDE model to generate conformations, followed by a 3D GNN for property prediction.
}
\label{fig:downstream_pipeline}
\vspace{-2ex}
\end{figure}

\begin{table}[htb!]
\centering
\vspace{-1ex}
\caption{
\small
Results for molecular property prediction with \textbf{SchNet} as 
backbone. 
The geometric structures (conformers) are generated using either MMFF or \model{}.
We report the mean (and standard deviation) ROC-AUC of 3 seeds with scaffold splitting for each downstream task. CG denotes 
Conformation Generation. 
}
\label{tab:main_results_moleculenet_3D}
\begin{adjustbox}{max width=0.48\textwidth}
\small
\setlength{\tabcolsep}{4pt}
\centering
\begin{tabular}{l l c c c c c c c c c}
\toprule
Model & CG Method & BBBP $\uparrow$ & Sider $\uparrow$ & ClinTox $\uparrow$ & Bace $\uparrow$\\
\midrule
GIN & -- & 64.1$\pm$1.79 & 58.4$\pm$0.50 & 63.1$\pm$7.21 & 76.5$\pm$2.96 \\
\midrule
SchNet & MMFF & 61.4$\pm$0.29  & 59.4$\pm$0.27  & 64.6$\pm$0.50  & 74.3$\pm$0.66\\
SchNet & ConfGF & 62.7$\pm$1.97  & 60.1$\pm$0.87  & 64.1$\pm$2.83  & 73.2$\pm$3.53\\
SchNet & ClofNet & 61.7$\pm$1.19  & 56.0$\pm$0.10  & 58.2$\pm$0.44  & 62.5$\pm$0.17\\
SchNet & \model{} & \textbf{65.2$\pm$0.43} & \textbf{60.5$\pm$0.39} & \textbf{72.9$\pm$1.02} & \textbf{78.6$\pm$0.40}\\
\bottomrule
\end{tabular}
\end{adjustbox}
\vspace{-3ex}
\end{table}

\subsection{Downstream with Topology to Conformation} \label{sec:downstream_2Dto3D}
We note that the pretraining in \model{} does two things: representation learning and topology/conformation generation. Such a conformation generation pretraining enables more prosperous downstream tasks. Here we consider 4 MoleculeNet tasks where only the 2D molecular graphs are available. Then we apply the SE(3)-equivariant conformation generation in \model{}, after which a 3D GNN will be trained for property prediction (\Cref{fig:downstream_pipeline}).

We consider three classic and state-of-the-art topology-to-conformation generation baselines. Merck molecular force field (MMFF)~\cite{halgren1996merck} is a heuristic method using physical simulation. ConfGF~\cite{shi2021learning} is an SE(3)-invariant conformation generation method using score matching, and ClofNet~\cite{du2022se} is the state-of-the-art SE(3)-equivariant conformation generation method using GNN. With the generated conformations, we apply a SchNet model (without pretraining) for property prediction. We also add a 2D GNN baseline with the atom and bond type information.
As shown in~\Cref{tab:main_results_moleculenet_3D}, we can observe that the CG baselines act worse than the 2D GNN for property prediction. Yet, \model{} can beat both the 2D and CG baselines. More discussions are in~\Cref{sec:ablation_studies}.

\subsection{Discussion on \model{}}
\textbf{For pretraining, data reconstruction is stronger than latent representation reconstruction.}
Starting from BOYL~\cite{grill2020bootstrap} and SimSiam~\cite{chen2021exploring}, the non-contrastive SSL methods have been widely explored. GraphMVP~\cite{liu2021pre} summarizes that these methods essentially reconstruct the latent representation space. Our proposed \model{} further proves that directly applying the data reconstruction is superior on the graph data. This observation also aligns well in the vision domain~\cite{he2022masked}. Complete results can be found in~\Cref{sec:ablation_studies}.

\section{Conclusion and Outlook}
We proposed \model{}, a group symmetric pretraining method on the 2D topology and 3D geometry modalities of molecules. \model{} introduces the first  SE(3)-equivariant and reflection-antisymmetric SDE for the topology to conformation generation and also the first  SE(3)-invariant SDE for the conformation to topology generation for molecule representation learning.
We provide theoretical insights that \model{} obtains tighter MI estimation over previous works.
We also empirically verified that \model{} retains essential knowledge from both modalities, resulting in state-of-the-art performance on 26 out of 32 tasks compared to 17 competitive baselines.

We note that multi-modal pretraining has been widely explored in drug discovery, not only between the topologies and conformations ({\ie}, the chemical structures), but also between the natural language and chemical structures. This research track is not exclusive to our work, and we believe that this can be a promising direction in the future exploration of the foundation model for molecule discovery.

\section*{Acknowledgement}
\vspace{-1ex}
This project is supported by the Natural Sciences and Engineering Research Council (NSERC) Discovery Grant, the Canada CIFAR AI Chair Program, collaboration grants between Microsoft Research and Mila, Samsung Electronics Co., Ltd., Amazon Faculty Research Award, Tencent AI Lab Rhino-Bird Gift Fund and two NRC Collaborative R\&D Projects (AI4D-CORE-06, AI4D-CORE-08). This project was also partially funded by IVADO Fundamental Research Project grant PRF-2019-3583139727.
\bibliography{reference.bib}
\bibliographystyle{icml2023}


\newpage
\appendix
\onecolumn

\section{Comparison to Related Works}
In~\Cref{tab:comparison_related_works}, we provide a comprehensive overview of existing works on single-modal and multi-modal pretraining methods. We categorize the pretraining methods into generative and contrastive learning methods

\begin{table}[h]
\small
\caption{
Comparison between \model\, and existing graph SSL methods.
}
\vspace{-2ex}
\label{tab:comparison_related_works}
\center
\setlength{\tabcolsep}{10pt}
\begin{adjustbox}{max width=\textwidth}
\begin{tabular}{l c c c c c c}
\toprule
\multirow{3}{*}{Pre-training} & \multicolumn{2}{c}{2D Topology} & \multicolumn{2}{c}{3D Conformation} & \multicolumn{2}{c}{2D Topology and 3D Conformation}\\
\cmidrule(lr){2-3}
\cmidrule(lr){4-5}
\cmidrule(lr){6-7}
& Generative & Contrastive & Generative & Contrastive & Generative & Contrastive\\
\midrule

AttrMask~\citep{hu2019strategies,liu2019n} & \checkmark & - & - & - & - & - \\
InfoGraph~\citep{velivckovic2018deep,sun2019infograph} & - & \checkmark & - & - & - & - \\
ContexPred~\citep{hu2019strategies} & - & \checkmark & - & - & - & - \\
GraphCL~\citep{You2020GraphCL} & - & \checkmark & - & - & - & - \\
\midrule

Atom Type Prediction~\cite{liu2022molecular} & - & - & \checkmark & - & - & - \\
Distance Prediction~\cite{fang2021chemrl,liu2022molecular} & - & - & \checkmark & - & - & - \\
Angle Prediction~\cite{fang2021chemrl,liu2022molecular} & - & - & \checkmark & - & - & - \\
3D Infograph~\cite{liu2022molecular} & - & - & - & \checkmark & - & - \\
GeoSSL-RR Prediction~\cite{liu2022molecular} & - & - & \checkmark & - & - & - \\
GeoSSL-InfoNCE Prediction~\cite{liu2022molecular} & - & - & - & \checkmark & - & - \\
GeoSSL-EBM-NCE Prediction~\cite{liu2022molecular} & - & - & - & \checkmark & - & - \\
GeoSSL-DDM-1L~\cite{zaidi2022pre} & - & - & \checkmark & - & - & - \\
GeoSSL-DDM~\cite{liu2022molecular} & - & - & \checkmark & - & - & - \\

\midrule
3D InfoMax~\cite{stark20223d} & - & - & - & - & - & \checkmark \\
GraphMVP~\cite{liu2021pre} & - & - & - & - & \checkmark & \checkmark \\
GraphMVP-C~\cite{liu2021pre} & - & \checkmark & - & - & \checkmark & \checkmark \\
GraphMVP-G~\cite{liu2021pre} & \checkmark & - & - & - & \checkmark & \checkmark \\
\midrule
\model{} (ours) & - & - & - & - & \checkmark & \checkmark \\
\bottomrule
\end{tabular}
\end{adjustbox}
\end{table}

\section{Group Symmetry and Local Frame} \label{sec:app:group_symmetry}
\subsection{SE(3)/E(3) Group action and representations}
In this article, a 3D molecular graph is represented by a 3D point cloud. The corresponding symmetry group is $SE(3)$, which consists of translations and rotations.
Recall that we define the notion of equivariance functions in $\mathbf{R}^3$ in the main text through group actions. 
Formally, the group SE(3) is said to act on $\mathbf{R}^3$ if there is a mapping $\phi: SE(3) \times \mathbf{R}^3 \rightarrow \mathbf{R}^3$ satisfying the following two conditions:
\begin{enumerate}
    \item if $e \in SE(3)$ is the identity element, then
    $$\phi(e,\vr) = \vr\ \ \ \ \text{for}\ \ \forall \vr \in \mathbf{R}^3.$$
    \item if $g_1,g_2 \in SE(3)$, then
    $$\phi(g_1,\phi(g_2,\vr)) = \phi(g_1g_2, \vr)\ \ \ \ \text{for}\ \ \forall \vr \in \mathbf{R}^3.$$
\end{enumerate}
Then, there is a natural $SE(3)$ action on vectors $\vr$ in $\mathbf{R}^3$ by translating $\vr$ and rotating $\vr$ for multiple times. For $g \in SE(3)$ and $\vr \in \mathbf{R}^3$, we denote this action by $g\vr$. Once the notion of group action is defined, we say a function $f:\mathbf{R}^3 \rightarrow \mathbf{R}^3$ that transforms $\vr \in \mathbf{R}^3$ is equivariant if:
$$f(g\vr) = g f(\vr),\ \ \  \text{for}\ \ \forall\ \  \vr \in \mathbf{R}^3.$$
On the other hand, $f:\mathbf{R}^3 \rightarrow \mathbf{R}^1$ is invariant, if $f$ is independent of the group actions:
$$f(g\vr) = f(\vr),\ \ \  \text{for}\ \ \forall\ \  \vr \in \mathbf{R}^3.$$
For some scenarios, our problem is chiral sensitive. That is, after mirror reflecting a 3D molecule, the properties of the molecule may change dramatically. In these cases, it's crucial to include reflection transformations into consideration. More precisely, we say an $SE(3)$-equivariant function $f$ is \textbf{reflection-antisymmetric}, if:
\begin{equation}
f(\rho \vr) \neq f(\vr),
\end{equation}
for some reflections $\rho \in E(3)$.

\subsection{Equivariant Frames}
Frame is a popular terminology in science areas. In physics, frame is equivalent to a coordinate system. For example, we may assign a frame to all observers, although different observers may collect different data under different frames, the underlying physics law should be the same. In other words, denote the physics law by $f$, then $f$ should be an equivariant function. \textbf{Notice:} there are multiple ways to construct frames, and in this section, we introduce a frame with a set of orthonormal bases.

Since there are three orthogonal directions in $\mathbf{R}^3$, a frame in $\mathbf{R}^3$ consists of three orthogonal vectors:
$$F = (\ve_1,\ve_2,\ve_3).$$
Once equipped with a frame (coordinate system), we can project all geometric quantities to this frame. For example, an abstract vector $\vr \in \mathbf{R}^3$ can be written as $\vr = (r_1,r_2,r_3)$ under frame $F$, if:
$\vr = r_1 \ve_1 + r_2 \ve_2 + r_3 \ve_3.$
An equivariant frame further requires the three orthonormal vectors in $(\ve_1,\ve_2,\ve_3)$ to be equivariant. Intuitively, an equivariant frame will transform  according to the global rotation or translation of the whole system. Once equipped with an equivariant frame, we can project equivariant vectors into this frame:
\begin{equation} \label{projection}
\vr = \Tilde{r}_1 \ve_1 + \Tilde{r}_2 \ve_2 + \Tilde{r}_3 \ve_3.    
\end{equation}
We call the process of $\vr \rightarrow \Tilde{r}:= (\Tilde{r}_1,\Tilde{r}_2,\Tilde{r}_3)$ the \textbf{projection} operation. Since $\Tilde{r}_i = \ve_i \cdot \vr_i$ is expressed as an inner product between equivariant vectors, we know that $\Tilde{r}$ consists of scalars. 

In this article, we assign an equivariant frame to each edge. Therefore, we call them the local frames that reflect the local geometry around each chemical bond, following \cite{du2022se,du2023new}.  Consider node $i$ and one of its neighbors $j$ with positions $\vx_i$ and $\vx_j$, respectively. The orthonormal equivariant frame $\mathcal{F}_{ij} := (\ve^{ij}_1,\ve^{ij}_2,\ve^{ij}_3)$ is defined with respect to $\vx_i$ and $\vx_j$ as follows:
\begin{equation}
(\frac{\vx_i - \vx_j}{\norm{\vx_i - \vx_j}}, \frac{\vx_i \times \vx_j}{\norm{\vx_i \times \vx_j}},\frac{\vx_i - \vx_j}{\norm{\vx_i - \vx_j}} \times \frac{\vx_i \times \vx_j}{\norm{\vx_i \times \vx_j}}  ).
\end{equation}
Note that this frame is translation invariant if the system's mass is zero.

\paragraph{Reflection-AntiSymmetric}
Since we implement the cross product $\times$ for building the local frames,  the third vector in the frame is a pseudo-vector.
Then, the \textbf{projection} operation is not invariant under reflections (the inner product between a vector and a pseudo-vector change signs under reflection). Therefore, our model is able to discriminate two 3D geometries with different chirality.

Our local frames also enable us to output equivariant vectors by multiplying scalars $(v_1,v_2,v_3)$ with the frame: $\vv = v_1 \cdot \ve_1 + v_2 \cdot \ve_2 + v_3 \cdot \ve_3.$  It's easy to check that $\vv$ is a $SE(3)$ equivariant (reflection-antisymmetric) vector.

\section{Denoising Score Matching}

\subsection{Energy-Based Model (EBM)} \label{sec:app:EBM}
Energy-based model (EBM) is a powerful tool for modeling the data distribution. The formulation is:
\begin{equation} \label{eq:EBM_original}
p_\theta(\vx) = \frac{\exp(-E(\vx))}{A} = \frac{\exp(f(\vx))}{A},
\end{equation}
where the bottleneck is the intractable partition function $A = \int_\vx \exp(-E(\vx)) d\vx$. Recently, there have been big progress~\cite{song2021train} in solving such an intractable function, including contrastive divergence~\cite{hinton2002training}, score matching~\cite{hyvarinen2005estimation,song2020score}, and noise contrastive estimation~\citep{gutmann2010noise}.

\subsection{Score Matching} \label{sec:app:EBM_SM}
There exists a family of solutions called \textbf{score matching} (SM) to solve~\Cref{eq:EBM_original}.
The core idea of SM is that for the generative task, we do not need to directly estimate the density, but we just need to know the score or gradient of the data distribution, {\ie}, $\nabla_{\vx} \log p(\vx)$. Then a Markov chain Monte Carlo (MCMC) strategy can be adopted for data generation.

\paragraph{Score}
The score is defined as the gradient of log-likelihood w.r.t. the data $\vx$:
\begin{equation} \label{eq:app:EBM_score}
\begin{aligned}
s_\theta(\vx) = \nabla_\vx \log p_\theta(\vx) = -\nabla_\vx E(\vx) - \nabla_\vx \log A = -\nabla_\vx E(\vx) = \nabla_\vx f(\vx).
\end{aligned}
\end{equation}
Thus, by taking the score, {\ie}, the gradient w.r.t. the data, the partition function term will disappear since it is a constant. The SM transforms the density estimation problem into a score (gradient) matching problem: if the first-order gradient of function ($s_\theta(\vx)$) can match, then the learned model distribution with EBM is able to capture the data distribution precisely.

\paragraph{Explicit Score Matching (ESM)}
For training the model, originally, SM~\cite{hyvarinen2005estimation} applies the Fisher divergence to measure the discrepancy between data distribution and model distribution, terms explicit score matching (ESM):
\begin{equation} \label{eq:app:EBM_ESM}
\begin{aligned}
D_F( p_{\text{data}}(\vx) || p_\theta(\vx) )
& = \frac{1}{2} \mathbb{E}_{p_{\text{data}}} \| \nabla_\vx \log p_\theta(\vx) - \nabla_x \log p_{\text{data}}(\vx) \|_2^2\\
& = \frac{1}{2} \mathbb{E}_{p_{\text{data}}} \| s_\theta(\vx) - \nabla_\vx \log p_{\text{data}}(\vx) \|_2^2.
\end{aligned}
\end{equation}
The expectation w.r.t. $p_{\text{data}}(\vx)$ can be approximated using Monte Carlo sampling, yet the second term of~\Cref{eq:app:EBM_ESM} is intractable to compute since it needs to know $\nabla_x \log p_{\text{data}}(\vx)$. There are multiple solutions to this, including Implicit Score matching (ISM)~\cite{hyvarinen2005estimation} which rewrites~\Cref{eq:app:EBM_ESM} using integration by parts; the other appealing solution is the denoising score matching (DSM). Both are introduced below.

\paragraph{Implicit Score Matching (ISM)}
It is still impractical to calculate~\Cref{eq:app:EBM_ESM} due to the $\nabla_\vx \log p_{\text{data}}(\vx)$ term. Under certain conditions~\cite{hyvarinen2005estimation}, the Fisher divergence can be rewritten using integration by parts, and we can turn it to the implicit score matching (ISM), {\ie}, ESM to ISM:
\begin{equation} \label{eq:app:EBM_ISM}
\begin{aligned}
\mathbb{E}_{p_{\text{data}}} [\| \nabla_\vx \log p_{\text{data}}(\vx) - s_\theta(\vx) \|^2]
& = \mathbb{E}_{p_{\text{data}}} \Big[ \frac{1}{2} (\nabla_\vx E_\theta(\vx))^2  + \text{tr}(\nabla_\vx^2 E_\theta(\vx)) \Big] + C \\
& = \mathbb{E}_{p_{\text{data}}} \Big[ \frac{1}{2} (\| s_\theta(\vx) \|^2  + \text{tr}(\nabla_\vx s_\theta(\vx)) \Big] + C,
\end{aligned}
\end{equation}
where $C$ is the constant. The drawback of~\Cref{eq:app:EBM_ISM} is that it requires computing the Trace of Hessian. It is computationally expensive as the computation of the full second derivatives is quadratic in the dimensionality of data.

\paragraph{Denoising Score Matching (DSM)}
Along this line, denoising score matching (DSM)~\cite{vincent2011connection} proposes an elegant solution by connecting the SM with denoising autoencoder. It first perturbs the data with a noise distribution, {\ie}, $q_\sigma(\tilde \vx| \vx)$, and the goal is to use SM to approximate the $q_\sigma(\tilde \vx) = \mathbb{E}_{p_{\text{data}}(\vx)} [q_\sigma(\tilde \vx|\vx)]$ with a model distribution $p_\theta(\tilde \vx)$. 
DSM~\cite{vincent2011connection} then calculates the Fisher divergence between the perturbed data distribution and perturbed model distribution, which leads to the following equation: 
\begin{equation} \label{eq:appp:DSM}
\begin{aligned}
D_F( q_\sigma(\tilde \vx) || p_\theta(\tilde \vx) )
& = \frac{1}{2} \mathbb{E}_{q_\sigma(\tilde \vx)} \Big[ \| \nabla_{\tilde \vx} \log q_\sigma(\tilde \vx) - \nabla_{\tilde \vx} \log p_\theta(\tilde \vx) \|^2 \big]\\
& = \frac{1}{2} \mathbb{E}_{q_\sigma(\vx, \tilde \vx)} \big[ \| \nabla_{\tilde \vx} \log q_\sigma(\tilde \vx | \vx) - s_\theta(\tilde \vx) \|^2 \big] + C.
\end{aligned}
\end{equation}
The detailed derivation of~\Cref{eq:appp:DSM} can be found in~\Cref{sec:proof_of_DSM}.
This is an elegant solution because under the Gaussian kernel, {\ie}, $q_\sigma(\tilde \vx|\vx) = \mathcal{N}(\tilde \vx|\vx, \sigma^2 I)$, we can have an analytical solution to $\nabla_{\tilde \vx} \log q_\sigma(\tilde \vx|\vx) = \frac{1}{\sigma^2}(\vx - \tilde \vx)$. This is essentially a direction moving from $\tilde \vx$ back to $\vx$, and DSM makes the score to match it. Finally, the objective becomes:
\begin{equation} \label{eq:appp:DSM_Gaussian}
\begin{aligned}
\mathcal{L}_{\text{DSM}}
& \approx \frac{1}{2N} \sum_{i=1}^N \Big[ \| \frac{\tilde \vx_i - x_i}{\sigma^2} + s_\theta(\tilde \vx_i) \|^2 \Big],\\
& = \frac{1}{2} \mathbb{E}_{q_\sigma(\vx, \tilde \vx)} \Big[ \| \frac{\vx - \tilde \vx}{\sigma^2} - s_\theta(\tilde \vx) \|^2 \Big]\\
& =  \frac{1}{2} \mathbb{E}_{p_{\text{data}}(\vx)} \mathbb{E}_{q_\sigma(\tilde \vx|\vx)} \big[ \| \frac{\tilde \vx - \vx}{\sigma^2} + s_\theta(\tilde \vx) \|^2 \Big].
\end{aligned}
\end{equation}
Additionally, \cite{vincent2011connection} also proves that DSM is equivalent to ESM. Though there exists certain drawbacks~\cite{song2021train}, DSM serves as a promising tool to enable the SM family as a more applicable solution to EBM.

\paragraph{Noise Conditional Score Network (NCSN)}
Recently, \cite{song2019generative} finds that perturbing data with random Gaussian noise makes the data distribution more powerful than SM model. Thus it proposes Noise Conditional Score Network (NCSN) that can perturb the data using various levels of noise and estimates scores at all levels simultaneously.
More concretely, NCSN chooses the Gaussian kernel as noise distribution, {\ie}, $q_\sigma(\tilde \vx|\vx) = \mathcal{N}(\tilde \vx|\vx,\sigma^2 I)$. With $L$ levels of noises, it extends~\Cref{eq:appp:DSM_Gaussian} as the following new objective:
\begin{equation} \label{eq:app:NCSN_Gaussian}
\begin{aligned}
\ell(\theta; \sigma_i) & =  \frac{1}{2} \mathbb{E}_{p_{\text{data}}(\vx)} \mathbb{E}_{q_{\sigma_i}(\tilde \vx|\vx)} \Big[ \| \frac{\tilde \vx - \vx}{\sigma_i^2} + s_\theta(\tilde \vx) \|^2 \Big]\\
\mathcal{L}_{\text{NCSN}} & = \frac{1}{L} \sum_{i=1}^L \lambda(\sigma_i) \ell(\theta; \sigma_i),
\end{aligned}
\end{equation}
where $\lambda(\sigma_i) > 0$ is a coefficient function on $\sigma_i$.

\paragraph{Sampling for SM}
For the SM family (including ESM, ISM, DSM and NCSN), once we have the score, we can sample the data by a MCMC sampling method called Langevin dynamics:
\begin{equation}
\begin{aligned}
\tilde \vx_{t+1}
&= \tilde \vx_{t} + \frac{\epsilon^2}{2} \nabla_{x} \log p_\theta(\vx_t) + \epsilon z_{t}\\
&= \tilde \vx_{t} + \frac{\epsilon^2}{2} s_\theta(\vx_{t}) + \epsilon z_{t},
\end{aligned}
\end{equation}
where $z_t \sim \mathcal{N}(0, I)$.

\paragraph{Discussion}
Till now, the SM family has provided a unique solution for the generative task. It may seem likelihood-free, but recently, another track on the diffusion model found that indeed these two research lines can contribute to the same formulation~\cite{song2020score}. The only difference is that the diffusion model starts with a variational approximation perspective.

\newpage
\subsection{Proof of DSM} \label{sec:proof_of_DSM}
\textit{Proof of~\Cref{eq:appp:DSM}.}\\
First to put this into the ESM, we can have:
\begin{equation} \label{eq:DSM_Proof_01}
\begin{aligned}
D_F( q(\tilde \vx) || p_\theta(\tilde \vx) )
& = \frac{1}{2} \mathbb{E}_{q(\tilde \vx)} \| \nabla_{\tilde \vx} \log p_\theta(\tilde \vx) - \nabla_{\tilde \vx} \log q(\tilde \vx) \|_2^2\\
& = \frac{1}{2} \mathbb{E}_{q(\tilde \vx)} \Big[ \| \nabla_{\tilde \vx} \log p_\theta(\tilde \vx) \|^2 + 2 \cdot \langle \nabla_{\tilde \vx} \log p_\theta(\tilde \vx), \nabla_{\tilde \vx} \log q(\tilde \vx)\rangle + \| \nabla_{\tilde \vx} \log q(\tilde \vx) \|^2 \Big]\\
& = \frac{1}{2} \mathbb{E}_{q(\tilde \vx)} \Big[ \| s_\theta(\tilde \vx) \|^2 + 2 \cdot \langle s_\theta(\tilde \vx), \nabla_{\tilde \vx} \log q(\tilde \vx)\rangle \Big] + C_1 ,
\end{aligned}
\end{equation}
where $C_1 = \frac{1}{2} \mathbb{E}_{q(\tilde \vx)} \| \nabla_{\tilde \vx} \log q(\tilde \vx) \|^2$ is a constant and does not depend on the model parameter $\theta$.

Then let's take out the second term, and we can have following:
\begin{equation} \label{eq:DSM_Proof_02}
\begin{aligned}
& \mathbb{E}_{q(\tilde \vx)} \Big[ \langle \nabla_{\tilde \vx} \log q(\tilde \vx), s_\theta(\tilde \vx)\rangle \Big]\\
= & \int_{\tilde \vx} q(\tilde \vx) \langle \nabla_{\tilde \vx} \log q(\tilde \vx), s_\theta(\tilde \vx)\rangle d \tilde \vx\\
= & \int_{\tilde \vx} \langle \nabla_{\tilde \vx} q(\tilde \vx), s_\theta(\tilde \vx)\rangle d \tilde \vx\\
= & \int_{\tilde \vx} \langle \nabla_{\tilde \vx} \int_x q(\vx) \cdot q(\tilde \vx|x) dx, s_\theta(\tilde \vx)\rangle d \tilde \vx\\
= & \int_{\tilde \vx} \int_x q(\vx) \cdot \langle \nabla_{\tilde \vx} q(\tilde \vx|x), s_\theta(\tilde \vx)\rangle dx d \tilde \vx\\
= & \int_{\tilde \vx} \int_x q(\vx) \cdot \langle q(\tilde \vx|x) \nabla_{\tilde \vx} \log q(\tilde \vx|x), s_\theta(\tilde \vx)\rangle dx d \tilde \vx\\
= & \int_{\tilde \vx} \int_x q(\vx) q(\tilde \vx|x) \cdot \langle \nabla_{\tilde \vx} \log q(\tilde \vx|x), s_\theta(\tilde \vx)\rangle dx d \tilde \vx\\
= & \mathbb{E}_{q(x,\tilde \vx)} \Big[  \langle \nabla_{\tilde \vx} \log q(\tilde \vx|x), s_\theta(\tilde \vx)\rangle \Big]
\end{aligned}
\end{equation}

So we can put this back to~\Cref{eq:DSM_Proof_01} and let ESM be as:
\begin{equation} \label{eq:DSM_Proof_03}
\begin{aligned}
D_F( q(\tilde \vx) || p_\theta(\tilde \vx) )
& = \frac{1}{2} \mathbb{E}_{q(\tilde \vx)} \Big[ \| s_\theta(\tilde \vx) \|^2 + 2 \cdot \langle s_\theta(\tilde \vx), \nabla_{\tilde \vx} \log q(\tilde \vx)\rangle \Big] + C_1\\
& = \frac{1}{2} \mathbb{E}_{q(x,\tilde \vx)} \Big[ \| s_\theta(\tilde \vx) \|^2 \Big] + \mathbb{E}_{q(x,\tilde \vx)} \Big[ \langle \nabla_{\tilde \vx} \log q(\tilde \vx|x), s_\theta(\tilde \vx)\rangle \Big] + C_1.
\end{aligned}
\end{equation}
And then we can get the following equivalent objective by some special reconstruction:
\begin{equation} \label{eq:DSM_Proof_04}
\begin{aligned}
D_F( q(\tilde \vx) || p_\theta(\tilde \vx) )
& = \frac{1}{2} \mathbb{E}_{q(x,\tilde \vx)} \Big[ \| s_\theta(\tilde \vx) \|^2 \Big] + \mathbb{E}_{q(x,\tilde \vx)} \Big[ \langle \nabla_{\tilde \vx} \log q(\tilde \vx|x), s_\theta(\tilde \vx)\rangle \Big] + C_2 + \Delta \\
& = \frac{1}{2} \mathbb{E}_{q(x,\tilde \vx)} \Big[ \| s_\theta(\tilde \vx) - \nabla_{\tilde \vx} \log q(\tilde \vx|x) \|^2 \Big] + \Delta.
\end{aligned}
\end{equation}
where $C_2 = \frac{1}{2} \mathbb{E}_{q(\tilde \vx)} \| \nabla_{\tilde \vx} \log q(\tilde \vx|x) \|^2$ is a re-constructured constant.\\
\textit{End of proof.}

\section{Diffusion Model}
Another generative modeling track is the denoising diffusion probabilistic model (DDPM)~\cite{sohl2015deep,ho2020denoising}. The diffusion model is composed of two processes: a forward process that adds noise to the data and a backward process that does denoising to generate the true data. Below we give a brief summary of the Gaussian diffusion model introduced in~\cite{ho2020denoising}.

\subsection{Pipeline of Denoising Diffusion Probabilistic Model}
\paragraph{Forward process}
Give a data point from the real distribution $\vx_0 \sim q(\vx)$, the forward diffusion process is that we add small amount of Gaussian noise to the sample in $T$ steps, producing a sequence of noisy samples $\vx_1, \hdots, \vx_T$. The step sizes are controlled by a variance schedule $\{\beta_t \in (0, 1)\}_{t=1}^T$:
\begin{align}
q(\vx_t|\vx_{t-1}) & = \mathcal{N}(\vx_t; \sqrt{1-\beta_t} \vx_{t-1}, \beta_t I),\\
q(\vx_{1:T}|\vx_0) & = \prod_{t=1}^T q(\vx_t|\vx_{t-1}).
\end{align}

A nice property of the forward process is that we can sample $\vx_t$ at any arbitrary timestep $t$ in a closed form using the reparameterization trick. Let $\alpha_t = 1-\beta_t$ and $\bar \alpha_t = \prod_{i=1}^t \alpha_i$, we have:
\begin{equation} \label{eq:x_t_given_x_0}
\begin{aligned}
q(\vx_t|\vx_0) = \mathcal{N}(\vx_t; \sqrt{\bar \alpha_t} \vx_0, (1-\bar\alpha_t)I).
\end{aligned}
\end{equation}

Then using the Bayes theorem,  $q(\vx_{t-1} | \vx_t, \vx_0)$ can be written as a Guassian:
\begin{equation} \label{eq:reverse_posterior}
\begin{aligned}
q(\vx_{t-1} | \vx_t, \vx_0)
& = \mathcal{N}(\vx_{t-1}; \tilde \mu(\vx_t, \vx_0), \tilde \beta_t I)\\
& = \mathcal{N}(\vx_{t-1}; \frac{1}{\sqrt{\alpha_t}} \big( \vx_t - \frac{\beta_t}{\sqrt{1 - \bar \alpha_t}} \epsilon_t \big), \frac{1-\bar \alpha_{t-1}}{1 - \bar \alpha_t} \beta_t)
\end{aligned}
\end{equation}

By plugging in~\Cref{eq:x_t_given_x_0}, {\ie}, $\sqrt{\alpha_t} \vx_0 = \vx_t - \epsilon_t \sqrt{1-\bar \alpha_t}$, we can have
\begin{equation}
\begin{aligned}
\tilde \mu(\vx_t, \vx_0) 
& = \frac{\sqrt{\bar \alpha_{t-1}} (1-\alpha_t)}{1 - \bar \alpha_t} \vx_0 + \frac{\sqrt{\alpha_t} (1 - \bar \alpha_{t-1})}{1-\bar \alpha_t} \vx_t\\
& = \frac{(1-\alpha_t) (\vx_t - \epsilon_t \sqrt{1-\bar \alpha_t})}{(1 - \bar \alpha_t) \sqrt{\alpha_t}}  + \frac{\sqrt{\alpha_t} (1 - \bar \alpha_{t-1})}{1-\bar \alpha_t} \vx_t\\
& = \frac{1}{\sqrt{\alpha_t}} \vx_t -\frac{1-\alpha_t}{\sqrt{1-\bar \alpha_t} \sqrt{\alpha_t}} \epsilon_t\\
& = \frac{1}{\sqrt{\alpha_t}} \Big( \vx_t -\frac{1-\alpha_t}{\sqrt{1-\bar \alpha_t}} \epsilon_t \Big).
\end{aligned}
\end{equation}

\paragraph{Reverse process}
Under a reasonable setting for $\beta_t$ and $T$~\cite{dhariwal2021diffusion}, the distribution $q(\vx_T)$ is nearly an isotropic Gaussian, and sampling $\vx_T$ is trivial. Then for the reverse process, we are interested in the $q(\vx_{t-1}|\vx_t)$. \citep{sohl2015deep} claims that as $t \to \infty$ and $\beta_t \to 0$, $q(\vx_{t-1}|\vx_t)$ approaches a diagonal Gaussian distribution. To this end, it is sufficient to train a neural network to predict a mean $\mu_\theta$ and a diagonal covariance matrix $\Sigma_\theta = \sigma_t^2 I$:
\begin{equation}
\begin{aligned}
p_\theta(\vx_{t-1} | \vx_t) = \mathcal{N}(\vx_{t-1}; \mu_\theta(\vx_t, t), \sigma_t^2 I).
\end{aligned}
\end{equation}

\paragraph{Parameterization and variational lower bound}
The hidden variables are $\vx_{1:T}$, and inference is to infer the latent variables, {\ie}, $p(\vx_{1:T}|\vx_0)$. Variational inference is to use $p(\vx_{1:T}|\vx_0)$ to estimate the true posterior $q(\vx_{1:T}|\vx_0)$. If we use $\vx_{1:T}$ as $z$, and $\vx_0$ as $\vx$, then it resembles to the VAE. Recall that
\begin{equation}
\begin{aligned}
KL(q(\vz|\vx) || p(\vz|\vx))
& = \mathbb{E}_{q(\vz|\vx)} \big[ \log \frac{q(\vz|\vx)}{p(\vz|\vx)} \big]\\
& = \log p(\vx) + \mathbb{E}_{q(\vz|\vx)} \big[ \log \frac{q(\vz|\vx)}{p(\vx,\vz)} \big].
\end{aligned}
\end{equation}
Thus, to maximize the log-likelihood of estimated data distribution, $\log p(\vx)$, is equivalent to maximizing the variational lower bound $\mathbb{E}_{q(\vz|\vx)} \big[ \log \frac{p(\vx,\vz)}{q(\vz|\vx)} \big] = \mathbb{E}_{q(\vz|\vx)}\big[ p(\vx|\vz) \big] - KL(q(\vz|\vx) || p(\vz))$, which is the summation of a reconstruction term and a KL-divergence term:
\begin{equation}
\mathbb{E}_{q(\vz|\vx)} \big[ \log \frac{p(\vx,\vz)}{q(\vz|\vx)} \big] \Longrightarrow -\text{MSE} \big( \vx, h(\vz) \big) + \frac{1}{2} \sum_j \big( 1 + (\log \sigma_j^2) - \mu_j^2 - \sigma_j^2  \big).
\end{equation}
For training, people usually minimize the negative of ELBO: $\text{MSE} \big( \vx, h(\vz) \big) - \frac{1}{2} \sum_j \big( 1 + (\log \sigma_j^2) - \mu_j^2 - \sigma_j^2  \big)$.

Similarly in the diffusion model, we have:
\begin{equation}
\begin{aligned}
KL(q(\vx_{1:T}|\vx_0) || p(\vx_{1:T}|\vx_0))
& = \log p(\vx_0) + \mathbb{E}_{q(\vx_{1:T}|\vx_0)} \big[ \log \frac{q(\vx_{1:T}|\vx_0)}{p(\vx_{0:T})} \big],
\end{aligned}
\end{equation}
and our goal becomes to maximize the variational lower bound (VLB). It is equivalent to minimizing $\mathcal{L}(\vx_0) = -\mathcal{L}_{VLB}$, as:
\begin{equation}
\begin{aligned}
\mathcal{L}(\vx_0)
& = \mathbb{E}_{q} \Big[\log \frac{ q(\vx_{1:T}|\vx_0) }{p_\theta(\vx_{0:T})} \Big]\\
& = \mathbb{E}_{q} \Big[\log \frac{ \prod_{t=1}^T q(\vx_T|\vx_{t-1}) }{ p_\theta(\vx_T) \prod_{t=1}^T p_\theta(\vx_{t-1}|\vx_T) } \Big]\\
& = \mathbb{E}_{q} \Big[-\log p_\theta(\vx_T) + \sum_{t=2}^T \log \frac{ q(\vx_t|\vx_{t-1}) }{ p_\theta(\vx_{t-1}|\vx_t) } + \log \frac{q(\vx_1|\vx_0)}{p_\theta(\vx_0|\vx_1)} \Big]\\
& = \mathbb{E}_{q} \Big[-\log p_\theta(\vx_T) + \sum_{t=2}^T \log \frac{ q(\vx_t|\vx_{t-1}, \vx_0) }{ p_\theta(\vx_{t-1}|\vx_t) } + \log \frac{q(\vx_1|\vx_0)}{p_\theta(\vx_0|\vx_1)} \Big]\\
& = \mathbb{E}_{q} \Big[-\log p_\theta(\vx_T) + \sum_{t=2}^T \log \frac{ q(\vx_{t-1}|\vx_t,\vx_0) \cdot q(\vx_t|\vx_0) }{ p_\theta(\vx_{t-1}|\vx_t) \cdot q(\vx_{t-1}|\vx_0)} + \log \frac{q(\vx_1|\vx_0)}{p_\theta(\vx_0|\vx_1)} \Big] ~~~~ \text{// Baye's rule}\\
& = \mathbb{E}_{q} \Big[-\log p_\theta(\vx_T) + \sum_{t=2}^T \log \frac{ q(\vx_{t-1}|\vx_t,\vx_0)}{ p_\theta(\vx_{t-1}|\vx_t)} + \sum_{t=2}^T \log \frac{q(\vx_t|\vx_0)}{q(\vx_{t-1}|\vx_0)} + \log \frac{q(\vx_1|\vx_0)}{p_\theta(\vx_0|\vx_1)} \Big]\\
& = \mathbb{E}_{q} \Big[-\log p_\theta(\vx_T) + \sum_{t=2}^T \log \frac{ q(\vx_{t-1}|\vx_t,\vx_0) }{ p_\theta(\vx_{t-1}|\vx_0) } + \log \frac{q(\vx_T|\vx_0)}{p_\theta(\vx_0|\vx_1)} \Big]\\
& = \mathbb{E}_{q} \Big[ \log \frac{q(\vx_T|\vx_0)}{p_\theta(\vx_T)} + \sum_{t=2}^T \log \frac{ q(\vx_{t-1}|\vx_0,\vx_0) }{ q(\vx_{t-1}|\vx_0) } - \log p_\theta(\vx_0|\vx_1) \Big]\\
& = KL[q(\vx_T|\vx_0) || p_\theta(\vx_T)] + \sum_{t=2}^T KL[q(\vx_{t-1}|\vx_t,\vx_0) || p_\theta(\vx_{t-1}|\vx_t)] - \mathbb{E}_q [\log p_\theta(\vx_0|\vx_1)]\\
& = \underbrace{KL[q(\vx_T|\vx_0) || p_\theta(\vx_T)]}_{\mathcal{L}_T} + \sum_{t=2}^T \underbrace{KL[q(\vx_{t-1}|\vx_t,\vx_0) || p_\theta(\vx_{t-1}|\vx_t)]}_{\mathcal{L}_{t-1}} - \underbrace{\mathbb{E}_q [\log p_\theta(\vx_0|\vx_1)]}_{\mathcal{L}_0}.
\end{aligned}
\end{equation}

For the loss term $\mathcal{L}_t$, according to~\Cref{eq:reverse_posterior}, we have:
\begin{equation} \label{eq:app:DDPM_L_t}
\begin{aligned}
\mathcal{L}_t
& = \mathbb{E}_{q(\vx_{t-1}|\vx_t,\vx_0)} \Big[ \log \frac{q(\vx_{t-1}|\vx_t,\vx_0)}{p_\theta(\vx_{t-1}|\vx_t)} \Big]\\
& = \mathbb{E}_{\vx_0,\epsilon} \Big[ -\frac{1}{2 \| \Sigma_\theta(\vx_t,t) \|^2} \| \tilde \mu_t(\vx_t,\vx_0) - \mu_\theta(\vx_t, t) \|^2 \Big]\\
& = \mathbb{E}_{\vx_0,\epsilon} \Big[ -\frac{1}{2 \| \Sigma_\theta \|^2} \Big\| \tilde \mu_t(\vx_t,\vx_0) - \mu_\theta(\vx_t, t) \Big\|^2 \Big]\\
& = \mathbb{E}_{\vx_0,\epsilon} \Big[ -\frac{1}{2 \| \Sigma_\theta \|^2} \cdot \frac{1}{\alpha_t} \cdot \Big\| \vx_t - \frac{\beta_t}{\sqrt{1 - \bar \alpha_t}} \epsilon_t - \vx_t + \frac{\beta_t}{\sqrt{1 - \bar \alpha_t}} \epsilon_\theta(\vx_t, t)  \Big\|^2 \Big]\\
& = \mathbb{E}_{\vx_0,\epsilon} \Big[ -\frac{\beta_t^2}{2 \alpha_t (1-\bar \alpha_t)\| \Sigma_\theta \|^2} \cdot \| \epsilon_t - \epsilon_\theta(\vx_t, t) \|^2 \Big]\\
& = \mathbb{E}_{\vx_0,\epsilon} \Big[ -\frac{\beta_t^2}{2 \alpha_t (1-\bar \alpha_t)\| \Sigma_\theta \|^2} \cdot \| \epsilon_t - \epsilon_\theta(\sqrt{\bar \alpha_t}\vx_0 + \sqrt{1-\bar \alpha_t}\epsilon_t, t) \|^2 \Big].
\end{aligned}
\end{equation}

\paragraph{Simplification}
There is a nice strategy proposed in~\cite{ho2020denoising}: the objective function in~\Cref{eq:app:DDPM_L_t} can be simplified by ignoring the weighting term:
\begin{equation}
\mathcal{L}_t^{simple} = \mathbb{E}_{\vx_0,\epsilon} \Big[\| \epsilon_t - \epsilon_\theta(\sqrt{\bar \alpha_t}\vx_0 + \sqrt{1-\bar \alpha_t}\epsilon_t, t) \|^2 \Big].
\end{equation}

\subsection{Important Tricks}
The DDPM~\cite{ho2020denoising} also adopts the following tricks in the training and inference.

\begin{itemize}[noitemsep,topsep=0pt]
    \item $\Sigma_\theta(\vx_T, t) = \sigma_t^2I$. Then DDPM empirically tests $\sigma_t^2 = \beta_t$ and $\sigma_t^2 = \tilde \beta_t = \frac{1 - \hat \alpha_{t-1}}{1 - \hat \alpha_t} \beta_t$. Both have similar results. And these two are the two extreme choices corresponding to the upper and lower bounds on reverse process entropy with coordinatewise unit variance~\cite{sohl2015deep}.
    \item The second trick is that we model $p_\theta(\vx_{t-1}|\vx_T) = \mathcal{N}(\vx_{t-1}; \mu_\theta(\vx_T, t); \sigma_t^2 I)$. Then the loss term becomes:
\begin{equation}
L_{t-1} = \mathbb{E}_q [\frac{1}{2\sigma_t^2} \| \tilde \mu(\vx_T, \vx_0) - \mu_\theta(\vx_T, t) \|^2 ]
\end{equation}
    So one straightforward way is to directly model the mean, {\ie}, to match $\mu_\theta$ and $\tilde \mu$.
    \item Meanwhile, during the diffusion process, we can have $\vx_0$ and $\vx_T$. Thus, we can have $\tilde \mu (\vx_T, \vx_0)$ as a function of of $(\vx_T, \epsilon)$ or $(\vx_0, \epsilon)$. In specific, we can write it as:
\begin{equation}
\begin{aligned}
L_{t-1} = \mathbb{E}_q [\frac{1}{2\sigma_t^2} \| \frac{1}{\sqrt{\alpha_t}} \big( \vx_T - \frac{\beta_t}{\sqrt{1 - \bar \alpha_t}}\epsilon \big) - \mu_\theta(\vx_T, t) \|^2 ]
\end{aligned}
\end{equation}
    Since $\vx_T$ can be obtained by $\vx_0$, then we may as well model $\mu_\theta(\vx_T, t) = \tilde \mu_t(\vx_T, \vx_0(\vx_T, \epsilon_t)) = \frac{1}{\sqrt{\alpha_t}} (\vx_T - \frac{\beta_t}{\sqrt{1-\bar \alpha_t}} \epsilon_\theta(\vx_T, t))$, and the objective function becomes:
\begin{equation}
\begin{aligned}
L_{t-1} 
& = \mathbb{E}_q [\frac{1}{2\sigma_t^2} \| \frac{1}{\sqrt{\alpha_t}} \big( \vx_T - \frac{\beta_t}{\sqrt{1 - \bar \alpha_t}}\epsilon \big) - \frac{1}{\sqrt{\alpha_t}} (\vx_T - \frac{\beta_t}{\sqrt{1-\bar \alpha_t}} \epsilon_\theta(\vx_T, t)) \|^2 ]\\
& = \mathbb{E}_q [\frac{\beta_t^2}{2\sigma_t^2 \alpha_t(1-\bar \alpha_t)} \| \epsilon - \epsilon_\theta(\vx_T, t) \|^2 ]
\end{aligned}
\end{equation}

\item Thus, during sampling, the mean is
\begin{equation}
\begin{aligned}
\mu_\theta(\vx_T, t) = \frac{1}{\sqrt{\alpha_t}} (\vx_T - \frac{\beta_t}{\sqrt{1-\bar \alpha_t}} \epsilon_\theta(\vx_T, t)),
\end{aligned}
\end{equation}
thus the sampling is obtained by:
\begin{equation}
\begin{aligned}
\vx_{t-1} 
& = \frac{1}{\sqrt{\alpha_t}} (\vx_t - \frac{\beta_t}{\sqrt{1-\bar \alpha_t}} \epsilon_\theta(\vx_t, t)) + \sigma_t \vz. \quad\quad \text{//DDPM's paper}
\end{aligned}
\end{equation}

\item Further, if we want to model the score, {\ie}, the $\log_{\tilde x} \log q(\tilde x | x)$, and then the score network defined here needs to have a shift~\citep{song2020score}:
\begin{equation}
\begin{aligned}
\log_{\tilde x} q(\tilde x | x_0)
& = \log_{\vx_t} q(\vx_t | \vx_0)\\
& = - \log_{\vx_T} \mathcal{N}(\vx_T; \sqrt{\bar \alpha_t}\vx_0, (1-\bar \alpha_t)I)\\
& = \frac{1}{\sqrt{\alpha_t}} (\vx_t + \beta_t s_\theta(\vx_t, t)) + \sqrt{\beta_t} \vz.
\end{aligned}
\end{equation}

\end{itemize}

\section{Stochastic Differential Equation} \label{sec:app:SDE}
A more recent work~\cite{song2020score} unifies the score matching and DDPM into a unified framework, the stochastic differential equation (SDE). First let's do a quick recap on the NCSN and DDPM.

\subsection{Review of NCSN and DDPM}
\subsubsection{NCSN}
The objective function is:
\begin{equation}
\begin{aligned}
\mathcal{L} 
& = \sum_{t=1}^T \sigma_t^2 \cdot \mathbb{E}_{p_{data}(\vx)} \mathbb{E}_{q_{\sigma_t}(\vx_t|\vx)} \Big[\Big\| s_\theta(\vx_t, \sigma_t) - \nabla_{\vx_t} \log q_{\sigma_t}(\vx_t | \vx) \Big\|^2 \Big]\\
& = \sum_{t=1}^T \sigma_t^2 \cdot \mathbb{E}_{p_{data}(\vx)} \mathbb{E}_{q_{\sigma_t}(\vx_t|\vx)} \Big[\Big\| s_\theta(\vx_t, \sigma_t) - \frac{\vx - \vx_t}{\sigma_t^2} \Big\|^2 \Big].
\end{aligned}
\end{equation}
There is no explicit definition of forward and backward, and the \textbf{sampling} is achieved by the Langevine dynamics:
\begin{equation}
\vx_t^m = \vx_t^{m-1} + \frac{\epsilon^2}{2} s_\theta(\vx_t^{m-1}, \sigma_t) + \epsilon \vz_m, \quad\quad m=1, \hdots, M,
\end{equation}
where $\epsilon$ is the step size and $\vz_m \sim \mathcal{N}(0,I)$. The above is repeated with:
\begin{itemize}[noitemsep,topsep=0pt]
    \item From $t=T$ to $t=1$.
    \item $\vx_t^0 \sim \mathcal{N}(0, \sigma_t^2 I)$ when $t=T$.
    \item $\vx_t^0 = \vx_{t+1}^M$ when $t<T$.
\end{itemize}

\subsubsection{DDPM-$\epsilon$}
The \textbf{forward} is:
\begin{equation}
p(\vx_t|\vx_{t-1}) = \mathcal{N}(\vx_t; \sqrt{\alpha_t} \vx_{t-1}, (1-\alpha_t) I).
\end{equation}

The \textbf{backward} is:
\begin{equation}
\begin{aligned}
p_\theta(\vx_{t-1} | \vx_t) 
& = \mathcal{N}(\vx_{t-1}; \mu_\theta(\vx_t, t), \sigma_t^2 I)\\
& = \mathcal{N} \Big( \frac{1}{\sqrt{\alpha_t}} \Big( \vx_t - \frac{1 - \alpha_t}{\sqrt{1 - \bar \alpha_t}}  \epsilon_\theta(\vx_t, t) \Big), \sigma_t^2 I \Big),
\end{aligned}
\end{equation}
where $\sigma_t^2 = \beta_t$ or $\sigma_t^2 = \tilde \beta_t$.

The \textbf{objective} function is:
\begin{equation}
\mathcal{L} = \mathbb{E}_{\vx_0,\epsilon_t,t} \Big[\| \epsilon_t - \epsilon_\theta(\sqrt{\bar \alpha_t}\vx_0 + \sqrt{1-\bar \alpha_t}\epsilon_t, t) \|^2 \Big].
\end{equation}

The \textbf{sampling} is:
\begin{equation}
\begin{aligned}
\vx_{t-1} 
& = \mu_\theta(\vx_t, t) + \vz \sigma_t\\
& = \frac{1}{\sqrt{\alpha_t}} \Big( \vx_t - \frac{1 - \alpha_t}{\sqrt{1 - \bar \alpha_t}}  \epsilon_\theta(\vx_t, t) \Big) + \vz \sigma_t,
\end{aligned}
\end{equation}
where $\vz \sim \mathcal{N}(0, I)$. This is called ancestral sampling since it amounts to performing ancestral sampling from the graphical mode~\cite{song2020score}.

\subsubsection{DDPM-Score}
Equivalently, we can also model the score of DDPM, as will be discussed in the next subsection as well as in \cite{song2020score}. While it shares the same forward process as DDPM-$\epsilon$ introduced above, its backward process, objective, and sampling procedure differ, as outlined below.

The \textbf{backward} is:
\begin{equation}
p_\theta(\vx_{t-1} | \vx_t) = \mathcal{N}\Big(\vx_{t-1}; \frac{1}{\sqrt{\alpha_t}}(\vx_T + \beta_t s_\theta(\vx_T, t)), \beta_t I \Big),
\end{equation}

The \textbf{objective} function is:
\begin{equation}
\mathcal{L} = \mathbb{E}_{\vx_0,\vx_t, t} \Big[ \Big\| s_\theta(\vx_t, t) - \nabla_{\vx_t} \log q(\vx_t|\vx_0) \Big\|^2 \Big].
\end{equation}
The \textbf{sampling} is:
\begin{equation}
\vx_{t-1} = \frac{1}{\sqrt{\alpha_t}}(\vx_t + \beta_t s_\theta(\vx_t, t)) + \sqrt{\beta_t} \vz, \quad\quad t=T, \hdots, 1,
\end{equation}
where $\vz \sim \mathcal{N}(0, I)$.

\subsubsection{Comparison of $\epsilon$ and Score}
Note that in DDPM, it is modeling $KL(q(\vx_{t-1}|\vx_t,\vx_0) || p_\theta(\vx_{t-1}|\vx_t))$, while NCSN is modeling $s_\theta(\vx_t, t) - \nabla_{\vx_t} \log p(\vx_{t-1}|\vx_t)$ directly. Essentially, these two are equivalent, because:
\begin{equation}
\begin{aligned}
-\sqrt{1 - \bar \alpha_t} s_\theta(\vx_t, t) = \epsilon_\theta(\vx_t, t).
\end{aligned}
\end{equation}

\subsection{Stochastic Differential Equation}
Then we introduce how NCSN and DDPM are solutions to Stochastic Differential Equation (SDE). The SDE is also formulated using forward and backward processes.

The \textbf{forward} process is
\begin{equation}
d\vx = f(\vx, t) dt + g(t) dw,
\end{equation}
where $f(\vx,t)$ is the vector-value drift coefficient, $g(t)$ is the diffusion coefficient, and $dw$ is the Wiener process.

The \textbf{backward} process, as a remarkable result from\citep{anderson1982reverse}, is:
\begin{equation}
d\vx = [f(\vx,t) - g(t)^2 \nabla_\vx \log p(\vx_t)] dt + g(t) dw.
\end{equation}
Then the question is how to estimate the score $\nabla_\vx \log p_t(\vx)$.

\subsection{Objective Function of Diffusion Process}
According to~\cite{song2020score}, the objective of solutions to SDE can be written in the form of score matching:
\begin{equation}
\mathcal{L} = \mathbb{E}_t \mathbb{E}_{\vx_0} \mathbb{E}_{\vx_t | \vx_0} \Big[ \lambda(t) \Big\| s_\theta(\vx_t, t) - \nabla_{\vx_t} \log p(\vx_t | \vx_0) \Big\|^2 \Big],
\end{equation}
where $\lambda(t)$ is a weighting function. With sufficient data and model capacity, the optimal $s_\theta(\vx_t, t)$ equals to $\nabla_x \log p(\vx_t)$ for almost all $\vx_T$ and $t$. Then we will review how to match the DSM/NCSN and DDPM into this framework.

\subsection{Instantiations}
\paragraph{DSM/NCSN and VE SDE}
The discretization of VE SDE yields DSM/NCSN. The \textbf{forward} process is:
\begin{equation}
\begin{aligned}
\vx_t & = \vx_{t-1} + \sqrt{\sigma_t^2 - \sigma_{t-1}^2} \epsilon_{t-1}, &&\text{// discrete}\\
dx & = \sqrt{\frac{d[\sigma^2(t)]}{dt}}dw.  &&\text{// continuous}\\
\end{aligned}
\end{equation}

Assumption: the $\{ \sigma_{t} \}, t=1, 2, \hdots, T$ is a geometric sequence. Then the transition kernel becomes:
\begin{equation}
\begin{aligned}
dx = \sigma_{min} \Big( \frac{\sigma_{max}}{\sigma_{min}} \Big)^2 \sqrt{2 \log \frac{\sigma_{max}}{\sigma_{min}} } dw
\end{aligned}
\end{equation}
and the perturbation kernel can be obtained by
\begin{equation}
\begin{aligned}
p(\vx_t|\vx_0) = \mathcal{N}\Big(\vx_t; \vx_0, \sigma_{min}^2 \Big( \frac{\sigma_{max}}{\sigma_{min}} \Big)^{2t} I\Big)
\end{aligned}
\end{equation}

\paragraph{DDPM and VP SDE}
The forward process is:
\begin{equation}
\begin{aligned}
\vx_t & = \sqrt{\alpha_t} \vx_{t-1} + \sqrt{1-\alpha_t} \epsilon_{t-1}, &&\text{// discrete}\\\\
dx & = -\frac{1}{2} (1-\alpha_t) x dt + \sqrt{\beta(t)} dw.  &&\text{// continuous}
\end{aligned}
\end{equation}

If we use the arithmetic sequence for $\{\beta\}_{t=1}^T$, the transition kernel for VP SDE is:
\begin{equation}
\begin{aligned}
d\vx = -\frac{1}{2} (\beta_{\text{min}} + t(\beta_{\text{max}}-\beta_{\text{min}}) \vx dt + \sqrt{\beta_{\text{min}} + t(\beta_{\text{max}}-\beta_{\text{min}}) } dw
\end{aligned}
\end{equation}
and the perturbation kernel is:
\begin{equation}
\begin{aligned}
p(\vx_t | \vx_0) = \mathcal{N}\Big(\vx_t; e^{-\frac{1}{4}t^2 (\beta_{\text{max}} - \beta_{\text{min}} - \frac{1}{2}t \beta_{\text{min}}} \vx_0, I - I e^{-\frac{1}{2}t^2 (\beta_{\text{max}} - \beta_{\text{min}} - t \beta_{\text{min}}} \Big)
\end{aligned}
\end{equation}
The drift coefficient is $-\frac{1}{2} (\beta_{\text{min}} + t(\beta_{\text{max}}-\beta_{\text{min}}) x$, and the diffusion coefficient is $\beta_{\text{min}} + t(\beta_{\text{max}}-\beta_{\text{min}})$.

\newpage
\clearpage
\section{Mutual Information and Equivalent Conditional Likelihoods} \label{sec:app:MI_conditional_likelihoods}

To maximize the mutual information between variable $X, Y$ is equivalent to optimizing the following equation:
\begin{equation} \label{eq:app:MI_objective}
\mathcal{L} = \frac{1}{2} \mathbb{E}_{p(\vx,\vy)} \big[ \log p(\vy|\vx) + \log p(\vx|\vy) \big].
\end{equation}

We provide the proof for the discrete variable case below. For the continuous case, please check GeoSSL~\citep{liu2022molecular}.

\textit{Proof.}
First we can get a lower bound of MI. Assuming that there exist (possibly negative) constants $a$ and $b$ such that $a \le H(X)$ and $b \le H(Y)$, {\ie}, the lower bounds to the (differential) entropies, then we have:
\begin{equation}
\begin{aligned}
I(X;Y)
& = \frac{1}{2} \big( H(X) + H(Y) - H(Y|X) - H(X|Y) \big)\\
& \ge \frac{1}{2} \big( a + b - H(Y|X) - H(X|Y) \big)\\
& \ge \frac{1}{2} \big( a + b \big) + \mathcal{L},\\
\end{aligned}
\end{equation}
where the loss $\mathcal{L}$ is defined as:
\begin{equation}
\begin{aligned}
\mathcal{L} 
& = \frac{1}{2} \big( - H(Y|X) - H(X|Y) \big)\\ 
& = \frac{1}{2} \mathbb{E}_{p(\vx,\vy)} \Big[ \log p(\vx|\vy) \Big] + \frac{1}{2} \mathbb{E}_{p(\vx,\vy)} \Big[ \log p(\vy|\vx) \Big].
\end{aligned}
\end{equation}
\textit{End of proof.}

Empirically, we use energy-based models to model the distributions. The condition on the existence of $a$ and $b$ can be understood as the requirements that the two distributions ($p_\vx, p_\vy$) are not collapsed.

\subsection{Variational Representation Reconstruction} \label{sec:app:variational_representation_reconstruction}

The variational representation reconstruction (VRR) was first introduced in GraphMVP~\cite{liu2021pre}. There are two mirroring terms in~\Cref{eq:app:MI_objective}, and here we take one term for illustration. The goal of VRR is to take the variational lower bound to maximize:
\begin{equation} \label{eq:app:MI_objective_one_term}
\mathbb{E}_{p(\vx,\vy)} \big[ \log p(\vy|\vx) \big].
\end{equation}

The objective function to~\Cref{eq:app:MI_objective_one_term} has a variational lower bound as:
\begin{equation} \label{eq:variational_lower_bound}
\begin{aligned}
\log p(\vy|\vx)
\ge \mathbb{E}_{q(\vz_\vx|\vx)} \big[ \log p(\vy|\vz_\vx) \big] - KL(q(\vz_\vx|\vx) || p(\vz_\vx)). 
\end{aligned}
\end{equation}

And VRR proposes a proxy solution b doing the reconstruction on the representation space instead of the data space:
\begin{equation} \label{eq:variational_lower_bound_approximation}
\begin{aligned}
\mathcal{L}_{\text{G}} = \mathcal{L}_{\text{VRR}}
 = & \mathbb{E}_{q(\vz_\vx|\vx)} \big[ \| q_{\vx}(\vz_\vx) - \text{SG}(h_{\vy}) \|^2 \big] + \beta  KL(q(\vz_\vx|\vx) || p(\vz_\vx)).
\end{aligned}
\end{equation}

\newpage
\section{Implementation Details of \model{}} \label{sec:app:method}

In this section, we illustrate the details of our proposed \model{}, including the featurization, backbone models, hyperparameters, architectures for the score networks, etc.
The solution in~\Cref{sec:app:variational_representation_reconstruction} to~\Cref{eq:app:MI_objective} is indeed a conditional generative method to solving the self-supervised learning (SSL). Meanwhile, it is only a proxy solution by conducting the reconstruction on the representation space. Thus, we want to explore a more accurate and explicit estimation to the generative reconstruction on the data space ({\ie}, the 2D topology and 3D geometry of molecules).

\subsection{Backbone Models}
For the backbone models, we stick with the existing pretraining works~\cite{hu2019strategies,wang2022molecular,liu2021pre,liu2022molecular}, which can better illustrate the effectiveness of our proposed methods. We use Graph Isomorphism Network (GIN)~\citep{xu2018powerful} for modeling the 2D topology and SchNet~\cite{schutt2018schnet} for 3D conformation, respectively.

\subsection{Molecule Featurization} \label{sec:molecule_featurization}
The molecule featurization is an essential factor that should be taken into consideration. A recent work~\cite{sun2022rethinking} has empirically verified that utilizing the rich atom feature. We follow this strategy and employ the featurization from MoleculeNet~\cite{wu2018moleculenet} and OGB~\cite{hu2020ogb}. In specific, we have the atom and bond featurization in~\Cref{tab:molecule_featurization}.

\begin{table}[h]
\centering
\setlength{\tabcolsep}{5pt}
\fontsize{9}{9}\selectfont
\caption{
\small
Featurization for atoms and bonds.
}
\label{tab:molecule_featurization}
\begin{adjustbox}{max width=\textwidth}
\begin{tabular}{l l l}
\toprule
 & Hyperparameter & Value\\
\midrule
\multirow{11}{*}{Atom Featurization}
& Atom Type & [0, 118] \\
& Atom Chirality & \makecell[l]{\{unspecified, unrecognized type,\\~~~tetrahedral with clockwise rotation,\\~~~tetrahedral: counter-clockwise rotation\}} \\
& Atom Degree & [0, 10] \\
& Formal Charge & [-5, 5]\\
& Number of Hydrogen & [0, 8]\\
& Number of Unpaired Electrons & [0, 4] \\
& Hybridization & \{SP, SP2, SP3, SP3D, SP3D2\}\\
& Is Aromatic & \{False, True\} \\
& Is In Ring & \{False, True\} \\
\midrule
\multirow{3}{*}{Bond Featurization}
& Bond Type & \{single, double, triple, aromatic\} \\
& Bond Stereotype & \{none, Z variant, E variant, Cis, Trans, any\} \\
& Is conjugated & \{False, True\} \\
\bottomrule
\end{tabular}
\end{adjustbox}
\end{table}

We also want to highlight that such an atom featurization is only available for the topological graph; while for 3D conformation, only the atom type information is available. The other atom information requires either the topology information ({\eg}, degree, number of Hydrogen)  or chemical rules ({\eg}, chirality) to obtain, and they have not been utilized for the molecule geometric modeling~\cite{schutt2018schnet}.

\subsection{Pretraining Hyperparameters}
The pretraining pipeline is shown in~\Cref{fig:pretraining_pipeline} and the objective function is~\Cref{sec:ultimate_objective}. Below in~\Cref{tab:hyperparameter}, we illustrate the key hyperparameters used in \model{}.

\begin{table}[H]
\vspace{-4ex}
\centering
\setlength{\tabcolsep}{5pt}
\fontsize{9}{9}\selectfont
\caption{
\small
Hyperparameter specifications for \model{}.
}
\label{tab:hyperparameter}
\begin{adjustbox}{max width=\textwidth}
\begin{tabular}{l l}
\toprule
Hyperparameter & Value\\
\midrule
epochs & \{50, 100\} \\
learning rate 2D GNN & \{1e-5, 1e-6\} \\
learning rate 3D GNN & \{1e-5, 1e-6\} \\
SDE option & \{VE, VP\}\\
masking ratio $M$ & \{0, 0.3\}\\
$\beta$ & \{[0.1, 10]\} \\
number of steps & \{1000\}\\
$\alpha_1$ & \{0, 1\}\\
$\alpha_2$ & \{0\}\\
$\alpha_3$ & \{0\}\\
\bottomrule
\end{tabular}
\end{adjustbox}
\end{table}

\subsection{SE(3)-Equivariant SDE Model: From Topology to Conformation}

\begin{figure}[ht]
    \begin{subfigure}[\small 01 The general pipeline.]
    {\includegraphics[width=0.48\linewidth]{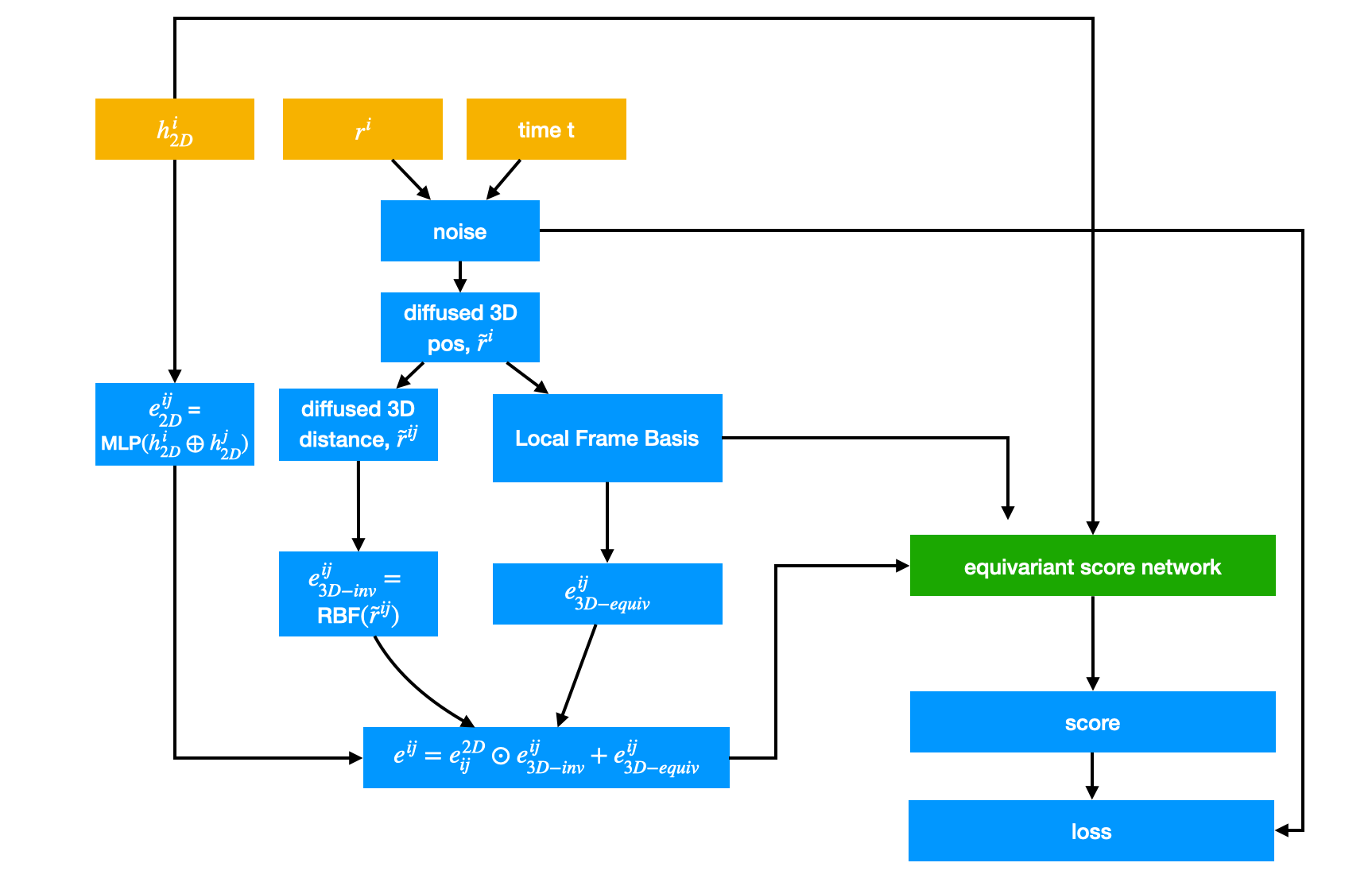}}
    \end{subfigure}
\hfill
     \begin{subfigure}[\small 02 The equivariant score network.]
     {\includegraphics[width=0.48\linewidth]{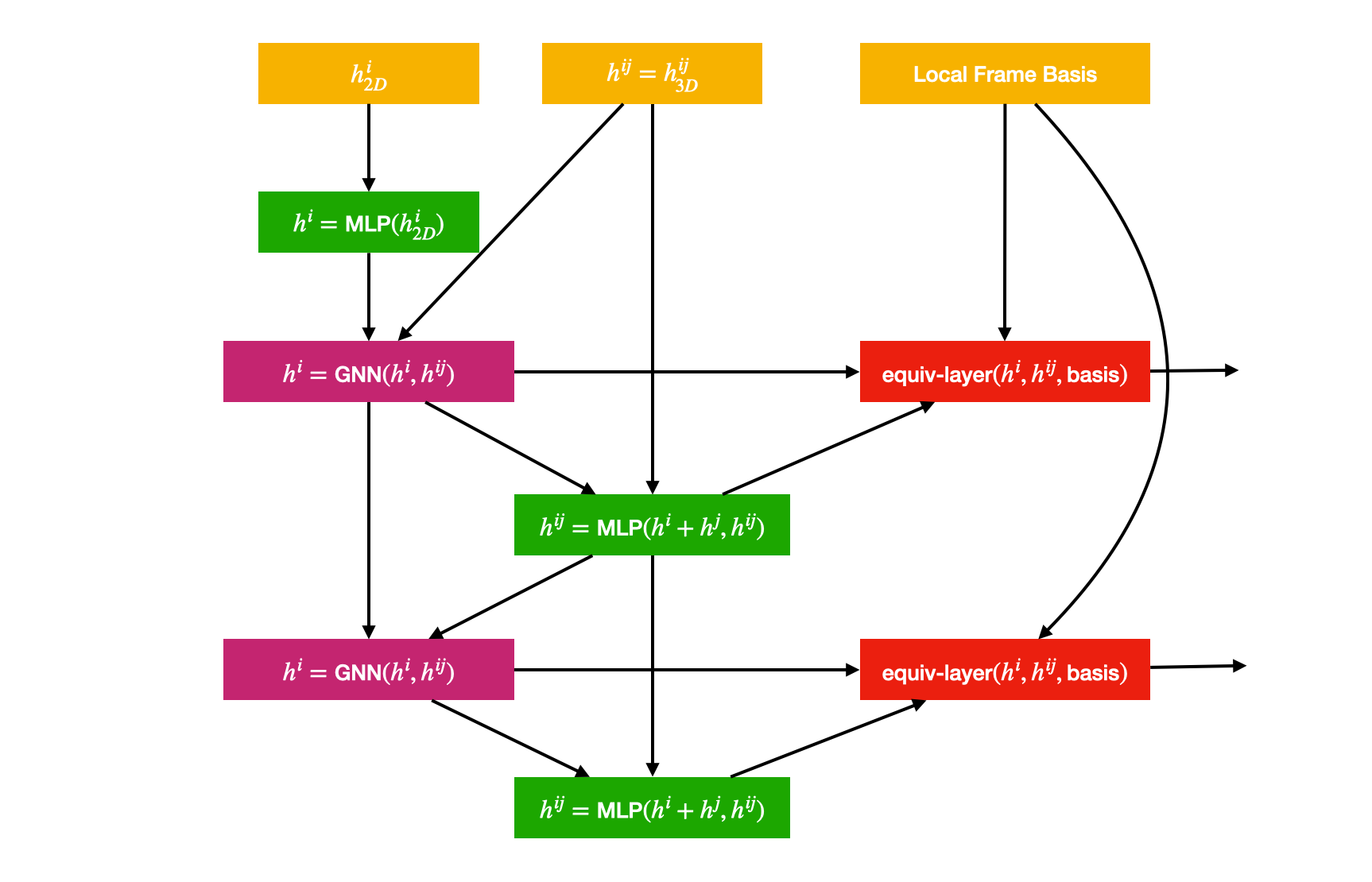}}
     \end{subfigure}
\\
     \begin{subfigure}[\small 03 The equivariant layer.]
     {\includegraphics[width=0.45\linewidth]{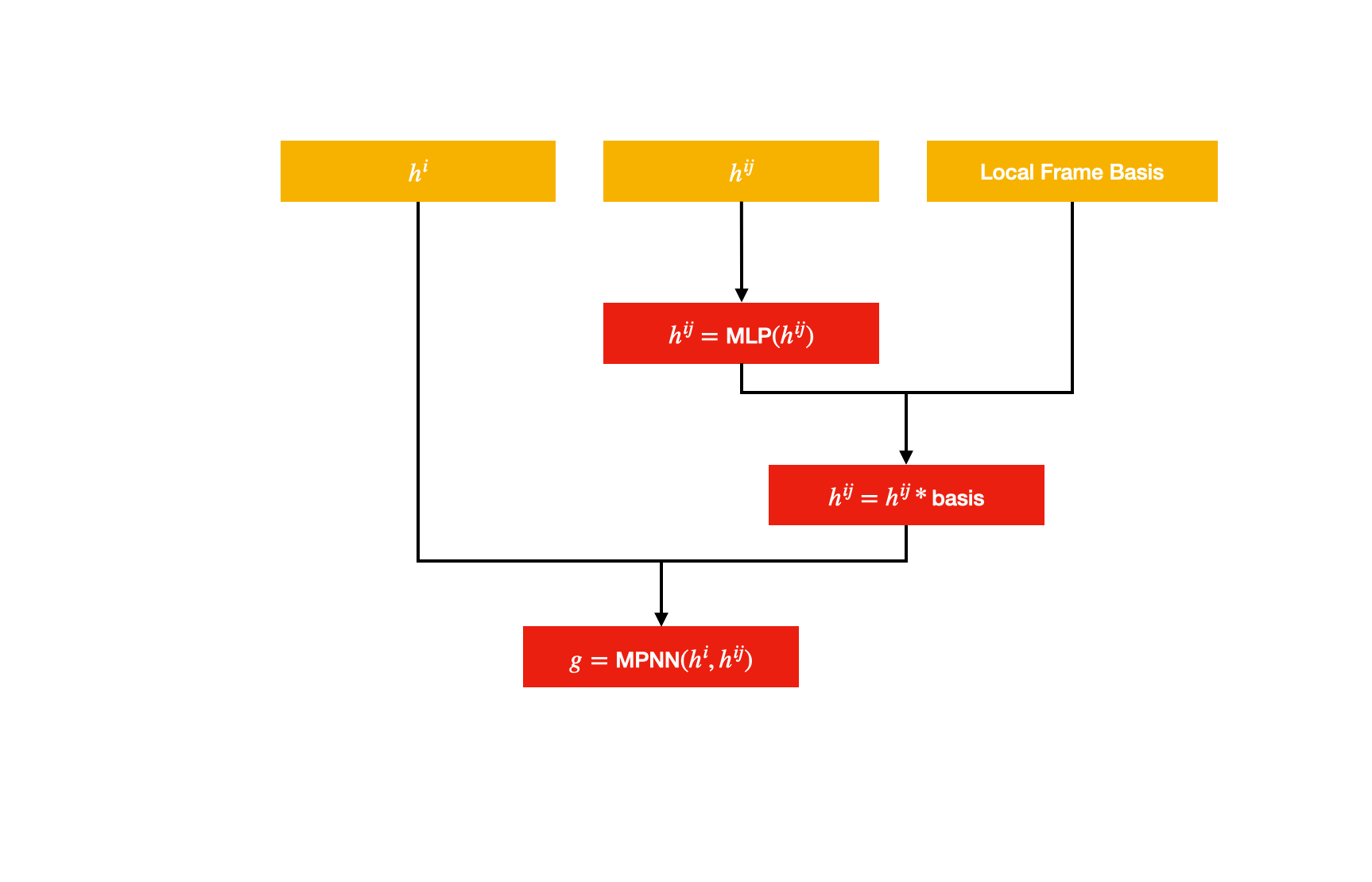}}
     \end{subfigure}
\vspace{-2.5ex}
\caption{
\small
Pipeline for the SE(3)-equivariant SDE model from topology to conformation.
}
\label{fig:pipeline_2D_to_3D}
\end{figure}

We list the detailed structure of the SE(3)-equivariant SDE model in~\Cref{fig:pipeline_2D_to_3D}.

\newpage
\subsection{SE(3)-Invariant SDE Model: From Conformation to Topology}
\begin{figure}[ht]
\centering
    \begin{subfigure}[\small 01 The general pipeline.]
    {\includegraphics[width=0.48\linewidth]{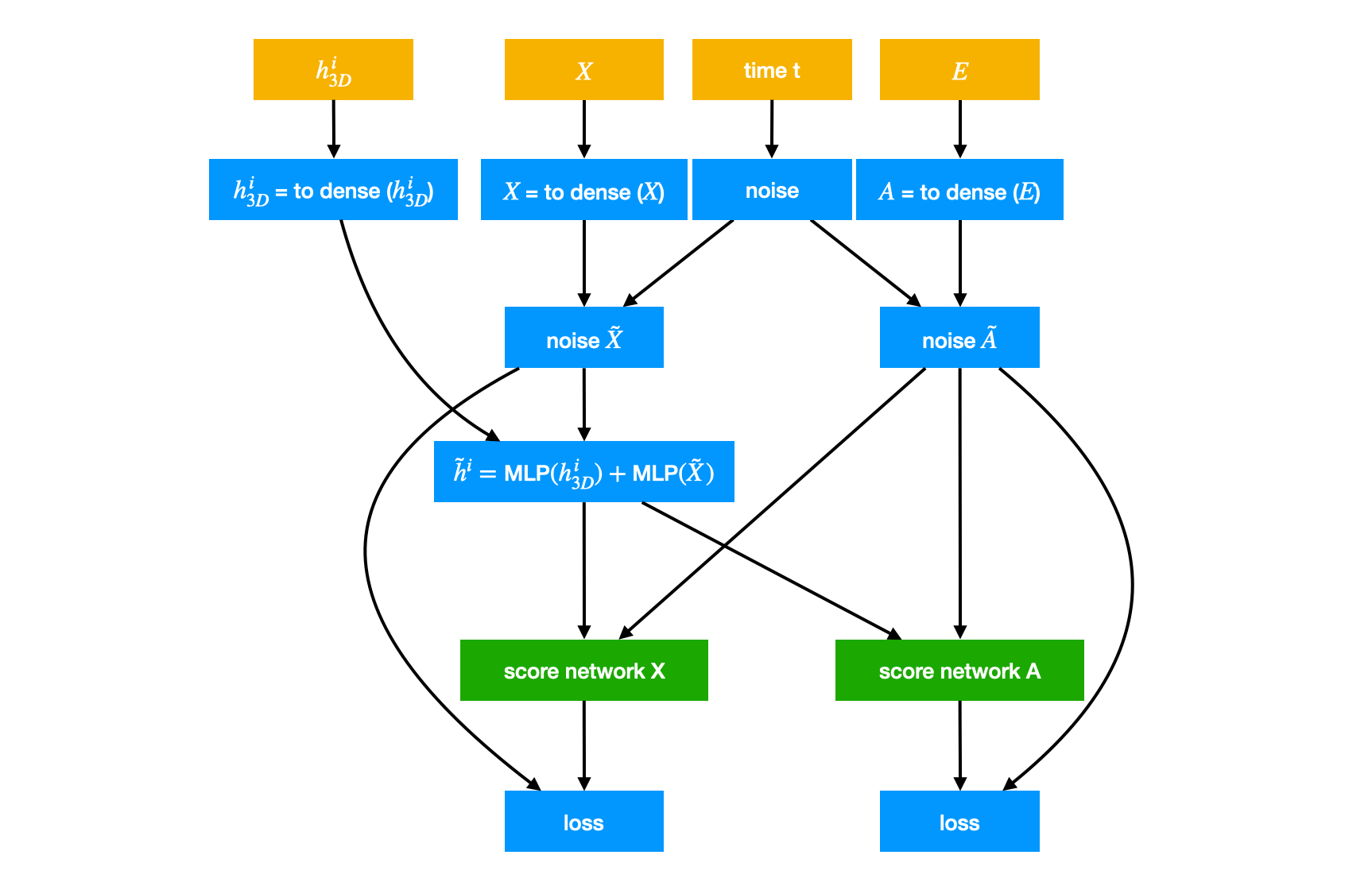}}
    \end{subfigure}
\hfill
     \begin{subfigure}[\small 02 The invariant score networks.]
     {\includegraphics[width=0.48\linewidth]{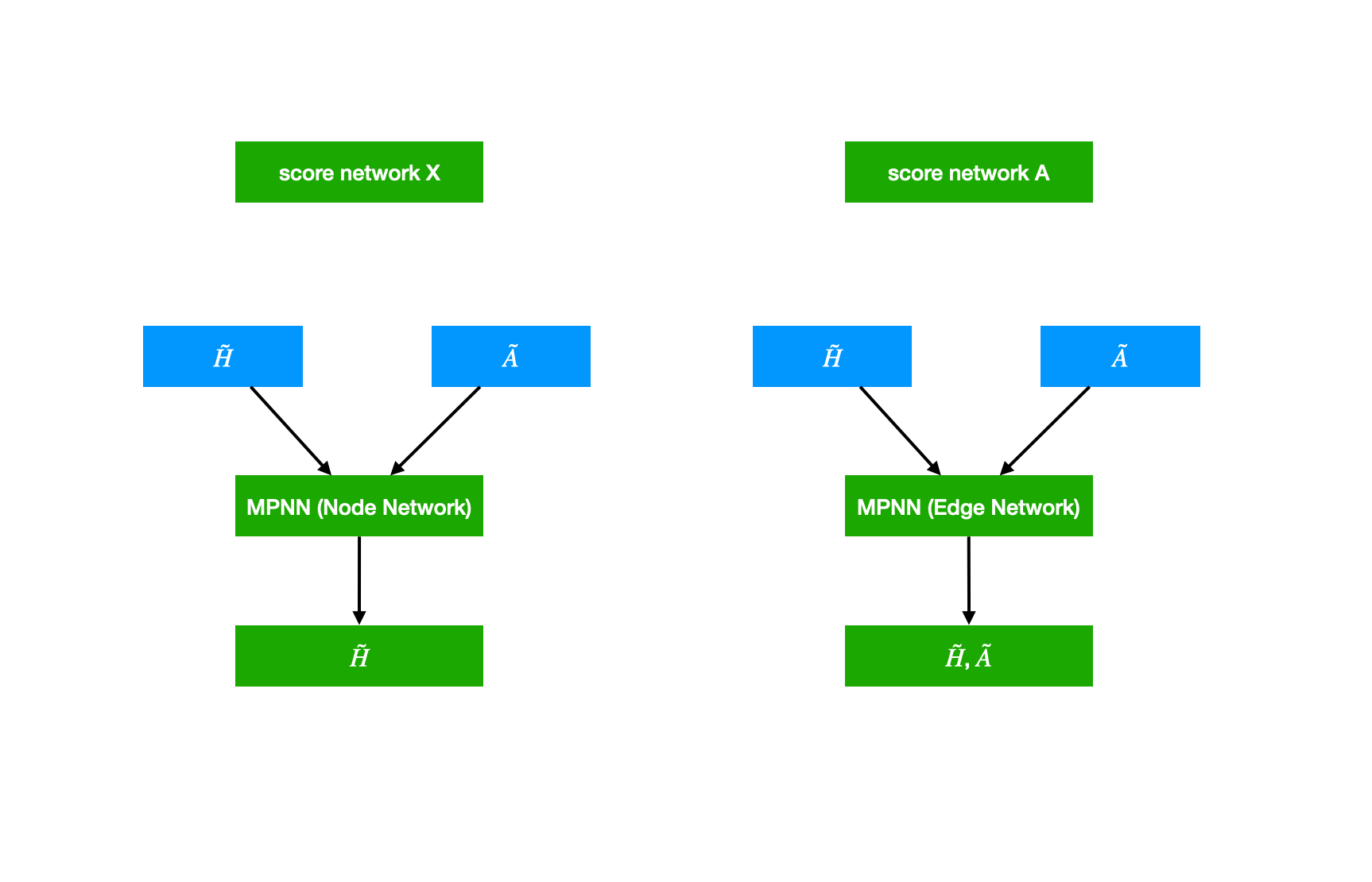}}
     \end{subfigure}
\vspace{-2.5ex}
\caption{
\small
Pipeline for the SE(3)-invariant SDE model from conformation to topology.
}
\label{fig:pipeline_3D_to_2D}
\vspace{-2ex}
\end{figure}

We list the detailed structure of the SE(3)-invariant SDE model in~\Cref{fig:pipeline_3D_to_2D}.

Similarly with the $2D \rightarrow 3D$ procedure, we first merge the 3D representation $\vy$ with the diffused atom feature ${\mX_t}$:
$$H_0 = \textbf{MLP}(\mX_t) + \vy.$$
Since the noised $\mE_t$ becomes a dense adjacency matrix, we will implement a densed GCN to model $\nabla_{\mX_t} \log p_t (\vx_t | \vy)$:
$$S_\theta^{\mX_t}(\vx_t) =  \textbf{MLP} (\text{Concate}\{H_0 || \cdots || H_L\}),$$
where $H_{i+1} = \textbf{GCN}(H_i,\mE_t)$. 
On the other hand, $\nabla_{\mE_t} \log p_t (\vx_t | \vy) \in \mathbf{R}^{n \times n}$ is modeled by an unnormalized dot product attention (without softmax):
$$S_\theta^{\mE_t}(\vx_t) = \textbf{MLP} (\{\text{Attention}(H_i)\}_{0 \leq i \leq L}).$$

\newpage
\clearpage
\section{Ablation Studies} \label{sec:ablation_studies}

This section provides more ablation studies to verify key concepts in molecule pretraining.

\subsection{Ablation Study on Generative SSL Pretraining}
We first provide a comprehensive comparison of the effect of the generative SSL part. 

\textbf{Pretraining.} In GraphMVP~\cite{liu2021pre}, the generative SSL is variational representation reconstruction (VRR). In \model{}, the generative SSL is composed of two SDE models.

\textbf{Downstream.} We consider both the 2D and 3D downstream tasks. We can tell that the generative SSL (SDE)) in \model{} is better than the generative SSL (VRR) in GraphMVP by a large margin on 27 out of 28 tasks.

\begin{table}[ht!]
\caption{
Ablation studies on generative SSL comparison.
Results for molecular property prediction tasks (with 2D topology only). 
The best results are marked in \underline{\textbf{bold}} and \textbf{bold}, respectively.
}
\centering
\begin{adjustbox}{max width=\textwidth}
\small
\setlength{\tabcolsep}{4pt}
\centering
\begin{tabular}{l c c c c c c c c c}
\toprule
Pre-training & BBBP $\uparrow$ & Tox21 $\uparrow$ & ToxCast $\uparrow$ & Sider $\uparrow$ & ClinTox $\uparrow$ & MUV $\uparrow$ & HIV $\uparrow$ & Bace $\uparrow$ & Avg $\uparrow$ \\
\midrule
VRR (GraphMVP) & 62.4$\pm$1.71 & 73.6$\pm$1.09 & 61.4$\pm$0.56 & 57.2$\pm$1.11 & \underline{\textbf{86.5$\pm$3.02}} & 75.5$\pm$1.58 & 75.4$\pm$0.96 & 72.7$\pm$2.16 & 70.61 \\
SDE-VE (\model{}) & \underline{\textbf{68.8$\pm$3.53}} & \underline{\textbf{76.5$\pm$0.28}} & \underline{\textbf{64.9$\pm$0.14}} & \textbf{59.2$\pm$0.44} & \textbf{86.1$\pm$2.15} & \textbf{77.7$\pm$2.15} & \textbf{77.0$\pm$0.66} & \underline{\textbf{79.6$\pm$0.66}} & \underline{\textbf{73.73}} \\

SDE-VP (\model{}) & \textbf{65.5$\pm$3.25} & \textbf{75.6$\pm$0.36} & \textbf{63.4$\pm$0.22} & \underline{\textbf{59.8$\pm$0.23}} & 81.1$\pm$1.83& \underline{\textbf{80.1$\pm$1.10}} & \underline{\textbf{78.6$\pm$0.31}} & \textbf{79.0$\pm$0.79} & \textbf{72.89} \\

\bottomrule
\end{tabular}
\end{adjustbox}
\vspace{-2ex}
\end{table}

\begin{table*}[htb]
\setlength{\tabcolsep}{5pt}
\fontsize{9}{9}\selectfont
\caption{
\small
Ablation studies on generative SSL comparison.
Results on 12 quantum mechanics prediction tasks from QM9. We take 110K for training, 10K for validation, and 11K for testing. The evaluation is mean absolute error.
The best results are marked in \underline{\textbf{bold}} and \textbf{bold}, respectively.
}
\begin{adjustbox}{max width=\textwidth}
\begin{tabular}{l c c c c c c c c c c c c}
\toprule
Pretraining & Alpha $\downarrow$ & Gap $\downarrow$ & HOMO$\downarrow$ & LUMO $\downarrow$ & Mu $\downarrow$ & Cv $\downarrow$ & G298 $\downarrow$ & H298 $\downarrow$ & R2 $\downarrow$ & U298 $\downarrow$ & U0 $\downarrow$ & Zpve $\downarrow$\\
\midrule

VRR (GraphMVP) & 0.058 & 44.64 & 27.32 & 22.50 & 0.030 & 0.030 & 14.96 & 14.69 & 0.127 & 14.35 & 13.96 & 1.680\\

SDE-VE (\model{})  & \underline{\textbf{0.056}} & \underline{\textbf{41.84}} & \underline{\textbf{25.79}} & \textbf{21.63} & \underline{\textbf{0.027}} & \underline{\textbf{0.029}} & \underline{\textbf{11.47}} & \underline{\textbf{10.71}} & \textbf{0.233} & \underline{\textbf{11.04}} & \underline{\textbf{10.95}} & \textbf{1.474}\\

SDE-VP (\model{}) & \textbf{0.056} & \textbf{42.75} & \textbf{25.84} & \underline{\textbf{21.52}} & \underline{\textbf{0.027}} & \underline{\textbf{0.029}} & \textbf{11.90} & \textbf{11.85} & \underline{\textbf{0.200}} & \textbf{12.03} & \textbf{11.69} & \underline{\textbf{1.453}}\\

\bottomrule
\end{tabular}
\end{adjustbox}
\vspace{-1ex}
\end{table*}

\begin{table*}[htb]
\setlength{\tabcolsep}{5pt}
\fontsize{9}{9}\selectfont
\centering
\caption{
\small
Ablation studies on generative SSL comparison.
Results on eight force prediction tasks from MD17. We take 1K for training, 1K for validation, and 48K to 991K molecules for the test concerning different tasks. The evaluation is the mean absolute error.
The best results are marked in \underline{\textbf{bold}} and \textbf{bold}, respectively.
}
\begin{adjustbox}{max width=\textwidth}
\begin{tabular}{l c c c c c c c c}
\toprule
Pretraining & Aspirin $\downarrow$ & Benzene $\downarrow$ & Ethanol $\downarrow$ & Malonaldehyde $\downarrow$ & Naphthalene $\downarrow$ & Salicylic $\downarrow$ & Toluene $\downarrow$ & Uracil $\downarrow$ \\
\midrule
VRR (GraphMVP) & \textbf{1.177} & 0.389 & 0.533 & \textbf{0.828} & 0.562 & 0.806 & \textbf{0.528} & 0.717\\
SDE-VE (\model{}) & 1.247 & \textbf{0.364} & \textbf{0.448} & \underline{\textbf{0.735}} & \underline{\textbf{0.483}} & \underline{\textbf{0.785}} & \underline{\textbf{0.480}} & \underline{\textbf{0.575}}\\
SDE-VP (\model{}) & \underline{\textbf{1.087}} & \underline{\textbf{0.358}} & \underline{\textbf{0.300}} & 0.880 & \textbf{0.517} & \textbf{0.788} & 0.540 & \textbf{0.675}\\
\bottomrule
\end{tabular}
\end{adjustbox}
\end{table*}

\newpage
\subsection{Ablation Study on Atom Features and Comparison with Conformation Generation Methods}
As recently discussed in~\cite{sun2022rethinking}, the atom feature plays an important role in molecule modeling, especially for 2D topological modeling. We carefully consider this in our work.

Note that for GIN in~\Cref{tab:main_results_moleculenet_2D}, we are using comprehensive atom and bond features, as shown in~\Cref{sec:molecule_featurization}.
For the ablation study in~\Cref{tab:main_results_moleculenet_3D}, to make it a fair comparison between GIN and SchNet, we further employed merely the atom type (the same as 3D conformation modeling) and the bond type for 2D topology modeling.
We name these two as "GIN with rich features" and the GIN in~\Cref{tab:main_results_moleculenet_2D} as "GIN with simple features", respectively.
The results and comparison with conformation generation methods are shown in~\Cref{tab:ablation_study_moleculenet_3D}.

\begin{table}[ht!]
\centering
\caption{
Ablation study on the effect of rich features for GIN and comparison with SchNet on conformation generation (CG) methods.
}
\label{tab:ablation_study_moleculenet_3D}
\begin{adjustbox}{max width=\textwidth}
\small
\centering
\begin{tabular}{l l c c c c c c c c c}
\toprule
Model & CG Method & BBBP & Sider & ClinTox & Bace\\
\midrule
GIN with rich features & -- & 68.1$\pm$0.59 & 57.0$\pm$1.33 & 83.7 $\pm$2.93 & 76.7$\pm$2.51 \\
GIN with simple features & -- & 64.1$\pm$1.79 & 58.4$\pm$0.50 & 63.1$\pm$7.21 & 76.5$\pm$2.96 \\
\midrule
SchNet & MMFF & 61.4$\pm$0.29  & 59.4$\pm$0.27  & 64.6$\pm$0.50  & 74.3$\pm$0.66\\
SchNet & ConfGF & 62.7$\pm$1.97  & 60.1$\pm$0.87  & 64.1$\pm$2.83  & 73.2$\pm$3.53\\
SchNet & ClofNet & 61.7$\pm$1.19  & 56.0$\pm$0.10  & 58.2$\pm$0.44  & 62.5$\pm$0.17\\
SchNet & \model{} & 65.2$\pm$0.43 & 60.5$\pm$0.39 & 72.9$\pm$1.02 & 78.6$\pm$0.40\\
\bottomrule
\end{tabular}
\end{adjustbox}
\end{table}

\textbf{Observation 1.}
We can tell that using rich or simple features plays an important role in GIN model.
This can be observed when comparing the GIN in~\Cref{tab:main_results_moleculenet_2D} and SchNet in~\Cref{tab:main_results_moleculenet_3D}, and we summarize them in the first two rows in~\Cref{tab:ablation_study_moleculenet_3D}.

\textbf{Observation 2.}
Additionally, we can tell that SchNet on \model{} can outperform GIN with simple features, showing that in terms of the \model{} can extract more useful geometric information. Meanwhile, GIN with rich features performs better on two tasks, especially a large margin in ClinTox. This reveals that the heuristic 2D topological information can also convey some information that is missing in \model{}.

Thus, the main message we want to deliver to the audience is that \model{} is better in terms of the conformation generation, and can be combined with the 2D topology modeling for future works.

\clearpage
\subsection{Ablation Study on The Effect of Contrastive Learning in \model{}}

\begin{table}[ht!]
\caption{
Ablation studies on $\alpha_1$ in \model{}.
Results for molecular property prediction tasks (with 2D topology only). 
The best results are marked in \textbf{bold} for each pair of $\alpha_1 \in \{0, 1\}$.
}
\centering
\begin{adjustbox}{max width=\textwidth}
\small
\setlength{\tabcolsep}{4pt}
\centering
\begin{tabular}{l l c c c c c c c c c}
\toprule
& $\alpha_1$ & BBBP $\uparrow$ & Tox21 $\uparrow$ & ToxCast $\uparrow$ & Sider $\uparrow$ & ClinTox $\uparrow$ & MUV $\uparrow$ & HIV $\uparrow$ & Bace $\uparrow$ & Avg $\uparrow$ \\
\midrule
\multirow{2}{*}{VE}
& 0 & 68.8$\pm$3.53& \textbf{76.5$\pm$0.28} & 64.9$\pm$0.14& 59.2$\pm$0.44& 86.1$\pm$2.15& 77.7$\pm$2.15& 77.0$\pm$0.66& 79.6$\pm$0.66& 73.73\\
& 1 & \textbf{73.2$\pm$0.48} & \textbf{76.5$\pm$0.33} & \textbf{65.2$\pm$0.31} & \textbf{59.6$\pm$0.82} & \textbf{86.6$\pm$3.73} & \textbf{79.9$\pm$0.19} & \textbf{78.5$\pm$0.28} & \textbf{80.4$\pm$0.92} & \textbf{74.98}\\
\midrule

\multirow{2}{*}{VP}
& 0 & 65.5$\pm$3.25& 75.6$\pm$0.36& 63.4$\pm$0.22& 59.8$\pm$0.23& 81.1$\pm$1.83& 80.1$\pm$1.10& 78.6$\pm$0.31& 79.0$\pm$0.79& 72.89\\
& 1 & \textbf{71.8$\pm$0.76} & \textbf{76.8$\pm$0.34} & \textbf{65.0$\pm$0.26} & \textbf{60.8$\pm$0.39} & \textbf{87.0$\pm$0.53} & \textbf{80.9$\pm$0.37} & \textbf{78.8$\pm$0.92} & \textbf{79.5$\pm$2.17} & \textbf{75.07}\\

\bottomrule
\end{tabular}
\end{adjustbox}
\vspace{-2ex}
\end{table}

\begin{table*}[htb]
\setlength{\tabcolsep}{5pt}
\fontsize{9}{9}\selectfont
\caption{
\small
Ablation studies on $\alpha_1$ in \model{}.
Results on 12 quantum mechanics prediction tasks from QM9. We take 110K for training, 10K for validation, and 11K for testing. The evaluation is the mean absolute error.
The best results are marked in \textbf{bold} for each pair of $\alpha_1 \in \{0, 1\}$.
}
\begin{adjustbox}{max width=\textwidth}
\begin{tabular}{l l c c c c c c c c c c c c}
\toprule
& $\alpha_1$ & Alpha $\downarrow$ & Gap $\downarrow$ & HOMO$\downarrow$ & LUMO $\downarrow$ & Mu $\downarrow$ & Cv $\downarrow$ & G298 $\downarrow$ & H298 $\downarrow$ & R2 $\downarrow$ & U298 $\downarrow$ & U0 $\downarrow$ & Zpve $\downarrow$\\
\midrule

\multirow{2}{*}{VE}
& 0 & 0.056& \textbf{41.84} & 25.79& 21.63& 0.027& \textbf{0.029} & \textbf{11.47} & \textbf{10.71} & 0.233& \textbf{11.04} & \textbf{10.95} & \textbf{1.474}\\

& 1 & \textbf{0.055} & 41.88 & \textbf{25.62} & \textbf{21.51} & \textbf{0.026} & \textbf{0.029} & 12.91 & 12.37 & \textbf{0.142} & 12.68 & 12.56 & 1.608\\

\midrule

\multirow{2}{*}{VP}
& 0 & 0.056& 42.75& 25.84& 21.52& 0.027& 0.029& \textbf{11.90} & \textbf{11.85} & 0.200& \textbf{12.03} & \textbf{11.69} & \textbf{1.453}\\

& 1 & \textbf{0.054} & \textbf{41.77} & \textbf{25.74} & \textbf{21.41} & \textbf{0.026} & \textbf{0.028} & 13.07 & 12.05 & \textbf{0.151} & 12.54 & 12.04 & 1.587\\

\bottomrule
\end{tabular}
\end{adjustbox}
\vspace{-1ex}
\end{table*}

\begin{table*}[htb]
\setlength{\tabcolsep}{5pt}
\fontsize{9}{9}\selectfont
\centering
\caption{
\small
Ablation studies on $\alpha_1$ in \model{}.
Results on eight force prediction tasks from MD17. We take 1K for training, 1K for validation, and 48K to 991K molecules for the test concerning different tasks. The evaluation is the mean absolute error.
The best results are marked in \textbf{bold} for each pair of $\alpha_1 \in \{0, 1\}$.
}
\begin{adjustbox}{max width=\textwidth}
\begin{tabular}{l l c c c c c c c c}
\toprule
& $\alpha_1$ & Aspirin $\downarrow$ & Benzene $\downarrow$ & Ethanol $\downarrow$ & Malonaldehyde $\downarrow$ & Naphthalene $\downarrow$ & Salicylic $\downarrow$ & Toluene $\downarrow$ & Uracil $\downarrow$ \\
\midrule
\multirow{2}{*}{VE}

& 0 & 1.247 & 0.364& 0.448& 0.735& 0.483& 0.785& \textbf{0.480} & 0.575\\
& 1 & \textbf{1.112} & \textbf{0.304} & \textbf{0.282} & \textbf{0.520} & \textbf{0.455} & \textbf{0.725} & 0.515& \textbf{0.447}\\
\midrule
\multirow{2}{*}{VP}
& 0 & \textbf{1.087} & 0.358& \textbf{0.300} & 0.880 & 0.517& 0.788& 0.540 & 0.675\\
& 1 & 1.244 & \textbf{0.315} & 0.338& \textbf{0.488} & \textbf{0.432} & \textbf{0.712} & \textbf{0.478} & \textbf{0.468}\\
\bottomrule
\end{tabular}
\end{adjustbox}
\end{table*}

\newpage
\subsection{PaiNN as Backbone}

\begin{table}[htb]
\centering
\setlength{\tabcolsep}{5pt}
\fontsize{9}{9}\selectfont
\caption{
\small
Results on 12 quantum mechanics prediction tasks from QM9, and the backbone model is PaiNN. We take 110K for training, 10K for validation, and 11K for testing. The evaluation is mean absolute error, and the best and the second best results are marked in \underline{\textbf{bold}} and \textbf{bold}, respectively.
}
\label{tab:main_QM9_PaiNN_result}
\begin{adjustbox}{max width=\textwidth}
\begin{tabular}{l c c c c c c c c c c c c}
\toprule
Pretraining & $\alpha$ $\downarrow$ & $\nabla \mathcal{E}$ $\downarrow$ & $\mathcal{E}_\text{HOMO}$ $\downarrow$ & $\mathcal{E}_\text{LUMO}$ $\downarrow$ & $\mu$ $\downarrow$ & $C_v$ $\downarrow$ & $G$ $\downarrow$ & $H$ $\downarrow$ & $R^2$ $\downarrow$ & $U$ $\downarrow$ & $U_0$ $\downarrow$ & ZPVE $\downarrow$\\
\midrule

-- & 0.049 & 42.73 & 24.46 & 20.16 & 0.016 & 0.025 & 8.43 & 7.88 & 0.169 & 8.18 & 7.63 & 1.419\\



Distance Prediction & 0.049 & 37.23 & 22.75 & 18.26 & 0.014 & 0.030 & 9.31 & 9.35 & 0.143 & 9.85 & 9.07 & 1.566\\

3D InfoGraph & 0.047 & 44.25 & 24.06 & 18.54 & 0.015 & 0.052 & 8.81 & 7.97 & 0.143 & 8.68 & 8.08 & 1.416\\

GeoSSL-RR & 0.046 & 41.20 & 23.93 & 19.36 & 0.016 & 0.025 & 8.32 & 8.17 & 0.174 & 7.99 & 8.20 & 1.438\\

GeoSSL-InfoNCE & 0.045 & 39.29 & 23.23 & 18.40 & 0.015 & 0.024 & 8.34 & 8.37 & \underline{\textbf{0.127}} & 7.45 & 8.34 & 1.356\\

GeoSSL-EBM-NCE & 0.045 & 38.87 & 22.71 & 17.89 & 0.014 & 0.082 & 8.28 & 7.35 & 0.130 & 7.85 & 7.68 & 1.338\\

3D InfoMax & 0.046 & 36.97 & 21.31 & 17.69 & 0.014 & 0.024 & 8.38 & 7.36 & 0.135 & 8.60 & 7.99 & 1.453\\

GraphMVP & 0.044 & 36.03 & 20.71 & 17.02 & 0.014 & 0.024 & 8.31 & 7.36 & 0.132 & 7.57 & 7.34 & 1.337\\

GeoSSL-DDM-1L & 0.045 & 36.13 & 20.59 & 17.26 & 0.014 & 0.024 & 9.45 & 8.43 & 0.128 & 8.88 & 8.16 & 1.380\\

GeoSSL-DDM & \textbf{0.043} & 35.55 & 20.57 & \textbf{16.95} & 0.014 & 0.024 & 8.25 & 7.42 & \underline{\textbf{0.127}} & 7.36 & 7.34 & 1.334\\

Uni-Mol & 0.277 & 40.56 & 21.25 & 23.99 & 0.014 & 0.039 & 9.16 & 9.14 & 0.340 & 9.31 & 8.59 & 1.433\\


\midrule

\model{} (VE) & 0.044 & \underline{\textbf{34.67}} & \underline{\textbf{20.14}} & 17.05 & \underline{\textbf{0.013}} & \underline{\textbf{0.023}} & \underline{\textbf{7.64}} & \textbf{7.05} & 0.139 & \underline{\textbf{6.88}} & \underline{\textbf{6.79}} & \underline{\textbf{1.273}}\\

\model{} (VP) & \underline{\textbf{0.042}} & \textbf{35.09} & \underline{\textbf{20.14}} & \underline{\textbf{16.78}} & \underline{\textbf{0.013}} & \underline{\textbf{0.023}} & \textbf{8.17} & \underline{\textbf{7.01}} & 0.133 & \textbf{7.30} & \textbf{7.05} & \textbf{1.315}\\

\bottomrule
\end{tabular}
\end{adjustbox}
\vspace{-1ex}
\end{table}

\begin{table}[htb]
\setlength{\tabcolsep}{5pt}
\fontsize{9}{9}\selectfont
\centering
\caption{
\small
Results on eight force prediction tasks from MD17, and the backbone model is PaiNN. We take 1K for training, 1K for validation, and 48K to 991K molecules for the test concerning different tasks. The evaluation is mean absolute error, and the best results are marked in \underline{\textbf{bold}} and \textbf{bold}, respectively.
}
\label{tab:main_MD17_PaiNN_result}
\begin{adjustbox}{max width=\textwidth}
\begin{tabular}{l c c c c c c c c}
\toprule
Pretraining & Aspirin $\downarrow$ & Benzene $\downarrow$ & Ethanol $\downarrow$ & Malonaldehyde $\downarrow$ & Naphthalene $\downarrow$ & Salicylic $\downarrow$ & Toluene $\downarrow$ & Uracil $\downarrow$ \\
\midrule

-- & 0.572 & 0.053 & 0.230 & 0.338 & 0.132 & 0.288 & 0.141 & 0.201\\



Distance Prediction & 0.480 & 0.053 & 0.200 & 0.296 & 0.131 & 0.265 & 0.171 & 0.168\\

3D InfoGraph & 0.554 & 0.067 & 0.249 & 0.353 & 0.177 & 0.331 & 0.179 & 0.213\\

GeoSSL-RR & 0.559 & 0.051 & 0.262 & 0.368 & 0.146 & 0.303 & 0.154 & 0.202\\

GeoSSL-InfoNCE & 0.428 & 0.051 & 0.197 & 0.337 & 0.127 & 0.247 & 0.136 & 0.169\\

GeoSSL-EBM-NCE & 0.435 & 0.048 & 0.198 & \textbf{0.295} & 0.143 & 0.245 & 0.132 & 0.172\\

3D InfoMax & 0.479 & 0.052 & 0.220 & 0.344 & 0.138 & 0.267 & 0.155 & 0.174\\

GraphMVP & 0.465 & 0.050 & 0.205 & 0.316 & \textbf{0.119} & 0.242 & 0.136 & 0.168\\

GeoSSL-DDM-1L & 0.436 & 0.048 & 0.209 & 0.320 & \textbf{0.119} & 0.249 & 0.132 & 0.177\\

GeoSSL-DDM & \textbf{0.427} & 0.047 & \underline{\textbf{0.188}} & 0.313 & 0.120 & \textbf{0.240} & \textbf{0.129} & 0.167\\

Uni-Mol & 0.487 & 0.048 & 0.217 & 0.329 & 0.151 & 0.299 & 0.141 & 0.182\\


\midrule

\model{} (VE) & \underline{\textbf{0.421}} & \underline{\textbf{0.043}} & 0.195 & \underline{\textbf{0.284}} & \underline{\textbf{0.105}} & \underline{\textbf{0.236}} & \underline{\textbf{0.123}} & \underline{\textbf{0.158}}\\

\model{} (VP) & 0.443 & \textbf{0.045} & \textbf{0.191} & 0.301 & 0.131 & 0.261 & 0.140 & \textbf{0.159}\\

\bottomrule
\end{tabular}
\end{adjustbox}
\vspace{-3ex}
\end{table}

\subsection{Experiments on Conformation Generation}

Notice that \model{} also supports the 2D to 3D conformation generation tasks, and we compare the following state-of-the-art models in~\Cref{tab:results_conformation_generation_GEOM}.

\begin{table}[htb!]
\caption{\small
Results on conformation generation without FF optimization. The datasets are GEOM QM9 and GEOM Drugs.}
\label{tab:results_conformation_generation_GEOM}
\centering
\begin{tabular}{l r r r r r r r r}
\toprule
\multirow{3}{*}{Methods}
& \multicolumn{4}{c}{GEOM QM9} & \multicolumn{4}{c}{GEOM Drugs}\\
\cmidrule(lr){2-5} \cmidrule(lr){6-9}
& \multicolumn{2}{c}{COV (\%) $\uparrow$} & \multicolumn{2}{c}{MAT ({\AA}) $\downarrow$}
& \multicolumn{2}{c}{COV (\%) $\uparrow$} & \multicolumn{2}{c}{MAT ({\AA}) $\downarrow$}\\
\cmidrule(lr){2-3} \cmidrule(lr){4-5} \cmidrule(lr){6-7} \cmidrule(lr){8-9}
& Mean & Median & Mean & Median & Mean & Median & Mean & Median\\
\midrule
RDKit & 83.26 & 90.78 & 0.3447 & 0.2935 & 60.91 & 65.70 & 1.2026 & 1.1252\\
ConfGF~\cite{shi2021learning} & 88.49 & 94.13 & 0.2673 & 0.2685 & 62.15 & 70.93 & 1.1629 & 1.1596\\
DMGC~\cite{zhu2022direct} & 96.23 & 99.26 & 0.2083 & 0.2014 & 96.52 & 100.00 & 0.7220 & 0.7161\\
RMCF-R~\cite{wang2022regularized} & -- & -- & -- & -- & 82.25 & 90.77 & 0.839 & 0.789\\
RMCF-C~\cite{wang2022regularized} & -- & -- & -- & -- & 87.12 & 96.26 & 0.749 & 0.709\\
\model{} (ours)
& 92.37 & 97.21 & 0.2423 & 0.2356  
& 85.42 & 99.49 & 0.9485 & 0.9041\\
\bottomrule
\end{tabular}
\end{table}

\newpage
\clearpage
\section{Computational Cost on Pretraining}

\begin{table}[htb!]
\caption{\small
Computational time on 17 pretraining algorithms.
All the jobs are running on one single V100 GPU card.
The four models in the first block are 2D SSL, the nine models in the second block are 3D SSL, and the four models in the last block are 2D-3D SSL.
}
\label{tab:my_label}
\centering
\begin{tabular}{l r}
\toprule

Pretraining Algorithm & min / epoch\\

\midrule

AttrMask & ~5.5 min/epoch \\
ContextPred & ~14 min/epoch\\
InfoGraph & ~6 min/epoch \\
MolCLR & ~10 min/epoch\\

\midrule

Type Prediction & ~7.75 min/epoch\\

Distance Prediction & ~6.7 min/epoch\\

Angle Prediction & ~8 min/epoch \\

3D InfoGraph & 7.5 min/epoch\\

GeoSSL-RR & ~9.7 min/epoch\\

GeoSSL-InfoNCE & ~10 min/epoch\\

GeoSSL-EBM-NCE & ~10.8 min/epoch\\

GeoSSL-DDM-1L & ~11.2 min/epoch\\

GeoSSL-DDM & ~18 min/epoch\\

\midrule

3D InfoMax & ~8.6 min/epoch\\

GraphMVP & ~11 min/epoch\\

\model{} (VE) & ~30 min/epoch\\

\model{} (VP) & ~30 min/epoch\\
\bottomrule
\end{tabular}
\end{table}

\end{document}